\documentclass[10pt,twocolumn,letterpaper]{article}

\usepackage{cvpr}      
\usepackage{wrapfig}
\usepackage{booktabs}
\usepackage[dvipsnames]{xcolor}
\usepackage{stmaryrd}
\usepackage{pifont}
\usepackage{layouts}

\usepackage[dvipsnames]{xcolor}

\definecolor{cvprblue}{rgb}{0.21,0.49,0.74}
\usepackage[pagebackref,breaklinks,colorlinks,citecolor=cvprblue]{hyperref}

\def\paperID{360} 
\def\confName{3DV\xspace}
\def\confYear{2024\xspace}

\title{SMORE: Simultaneous Map and Object REconstruction}

\author{Nathaniel Chodosh\thanks{Equal Contribution}\\
  Villanova University\\
  {\tt\small nchodosh@villanova.edu}
\and
Anish Madan$^*$\\
Carnegie Mellon University\\
 {\tt\small anishmad@cs.cmu.edu}
\and
Simon Lucey\\
University of Adelaide\\
{\tt\small simon.lucey@adelaide.edu.au}
\and
Deva Ramanan\\
Carnegie Mellon University\\
{\tt\small deva@cs.cmu.edu}
}

\begin{document}
\setlength{\tabcolsep}{2pt}
\maketitle

\begin{abstract}
We present a method for dynamic surface reconstruction of large-scale urban scenes from LiDAR. Depth-based reconstructions tend to focus on small-scale objects or large-scale SLAM reconstructions that treat moving objects as outliers. We take a holistic perspective and optimize a compositional model of a dynamic scene that decomposes the world into rigidly-moving objects and the background. To achieve this, we take inspiration from recent novel view synthesis methods and frame the reconstruction problem as a global optimization over neural surfaces, ego poses, and object poses, which minimizes the error between composed spacetime surfaces and input LiDAR scans. In contrast to view synthesis methods, which typically minimize 2D errors with gradient descent, we minimize a 3D point-to-surface error by coordinate descent, which we decompose into registration and surface reconstruction steps. Each step can be handled well by off-the-shelf methods without any re-training. We analyze the surface reconstruction step for rolling-shutter LiDARs, and show that deskewing operations common in continuous time SLAM can be applied to dynamic objects as well, improving results over prior art by an order of magnitude. Beyond pursuing dynamic reconstruction as a goal in and of itself, we propose that such a system can be used to auto-label partially annotated sequences and produce ground truth annotation for hard-to-label problems such as depth completion and scene flow. Please refer \url{https://anishmadan23.github.io/smore/} for more visual results.
\end{abstract}


\begin{figure*}
    \centering
    \frame{\includegraphics[width=0.45\textwidth]{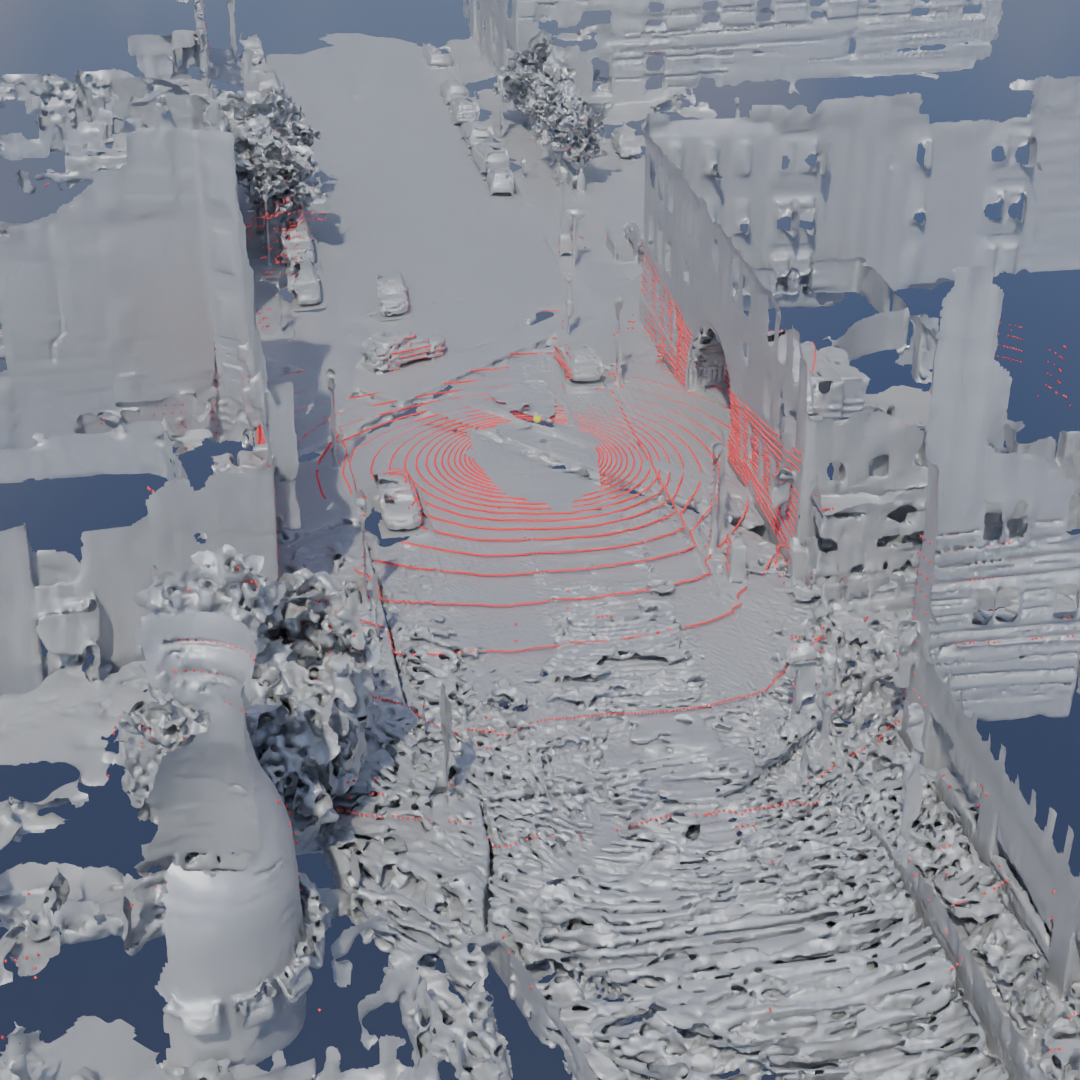}}
    \frame{\includegraphics[width=0.45\textwidth]{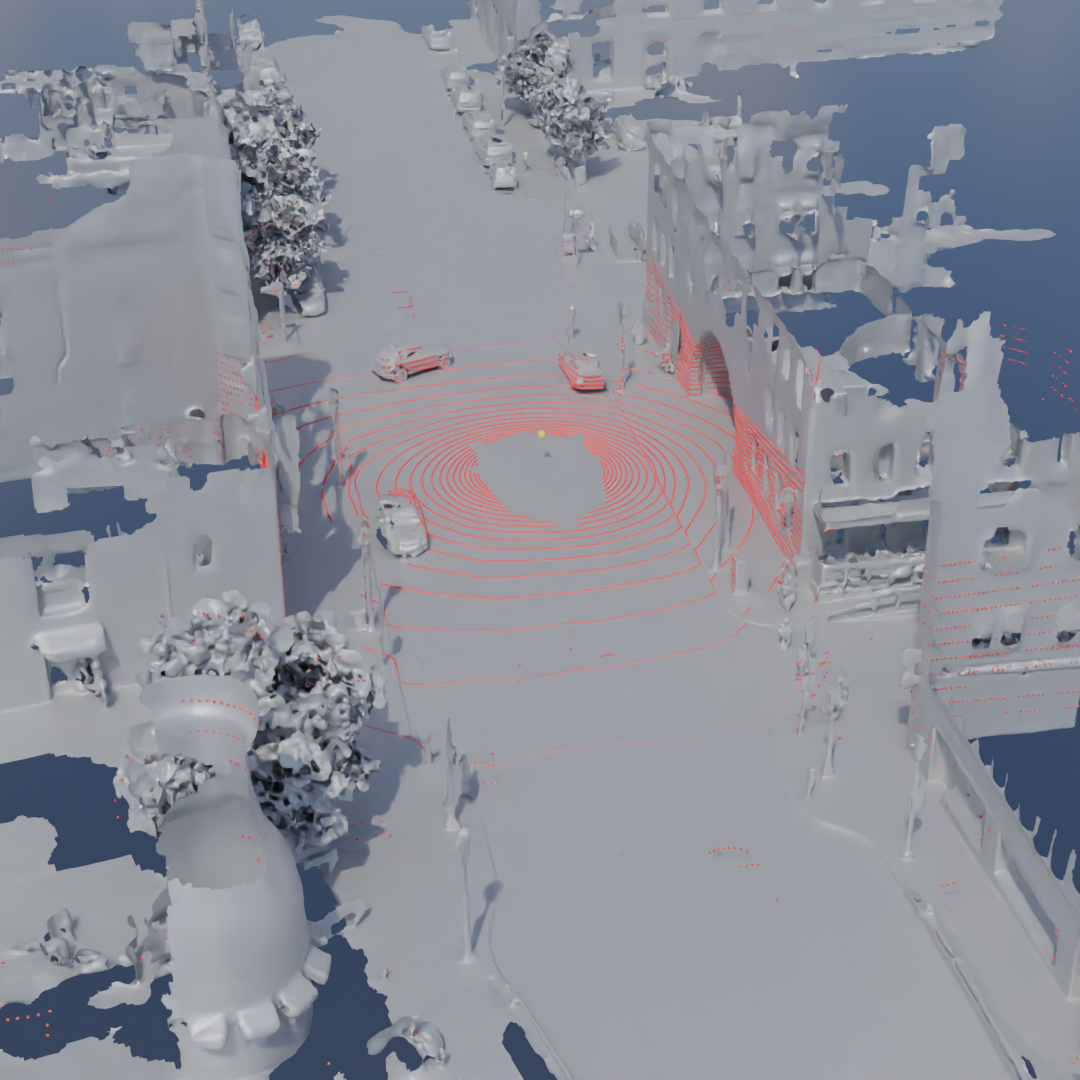}}\\
    \caption{NuScenes surface reconstruction produced by aggregating LiDAR scans using human-annotated ego-pose and dynamic object bounding boxes ({\bf left}). We introduce a global optimization that refines both ego and object poses so as to minimize a scan-to-surface reconstruction error, dramatically improving results ({\bf right}). To do so, we find it crucial to model rolling shutter LiDAR effects, particularly for dynamic objects (\cref{fig:ablation}). 
    Please see the animation in the supplement.\vspace{-1cm}}.
    \label{fig:title}
\end{figure*}

\section{Introduction}
\label{sec:intro}

\begin{figure}
    \centering
    \includegraphics[width=\linewidth]{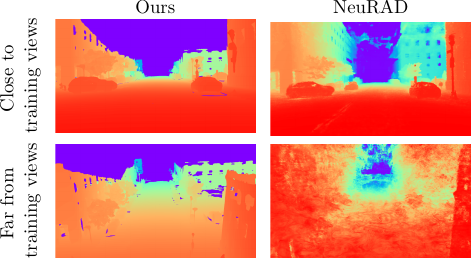}\\
    \caption{Depth maps produced by our method (\textbf{left}) as compared to those from a SOTA NeRF-based method\cite{tonderski2024neurad} (\textbf{right}). In the \textbf{top} row we see that when depth maps are close to the training views, \cite{tonderski2024neurad} and our method produce comparable results. However, in the \textbf{bottom} row, we see that moving the camera far from the training poses reveals large errors in the density field not present in our surface-based method.} 
    \label{fig:depth}
\end{figure}

Dynamic scene understanding aims to produce a model of the world that explains all measurements over time. In the context of depth sensors, this problem is posed as dynamic surface reconstruction, where the goal is to produce a time-varying surface that matches a sequence of depth measurements. This problem has been widely studied in the context of handheld RGB-D sensors capturing human-scale scenes~\cite{newcombe2015dynamicfusion, runz2018maskfusion, yu2021function4d, yang2022banmo}. However, investment in autonomous driving has created a new mode of depth capture --- spinning LiDAR sensors atop moving vehicles --- which is largely unaddressed by the existing research. Existing surface-based methods focus on reconstructing a few densely-scanned non-rigid objects, but autonomous driving scenes are typically composed of many sparsely-scanned rigid objects~\cite{gojcic2021weakly, chodosh2024}. In contrast, recent novel view synthesis methods have modeled AV scenes with a composition of rigid models~\cite{ost021nsg,ziyang2023snerf,turki2023suds,tonderski2024neurad}. However, we find that they produce poor-quality reconstructions of the underlying geometry (see \cref{fig:depth}). In this work, we aim to achieve the best of both worlds and present a compositional surface model of LiDAR sequences. We find that this approach produces superior estimates of world and actor geometry as well as superior pose estimates of the ego-vehicle and tracked objects.


{\bf Approach:} We address the dynamic scene reconstruction problem from a classic ``analysis by synthesis" perspective; we synthesize a spacetime reconstruction via a compositional model of geometry and motion. We then measure the 3D error of the reconstruction with respect to the observed LiDAR scans. Finally, we optimize the geometry and motion to minimize this 3D error. We take care to formulate the optimization so that it can be efficiently decomposed into alternating steps of 1) estimating 6-DOF motion parameters of rigidly-moving components (including the moving ego-vehicle) and 2) estimating the geometry of each rigid component (including the static background). Such a decomposition allows us to leverage off-the-shelf solutions to the point registration and point-to-surface reconstruction problems, respectively. However, a naive implementation does not produce high-quality results. Instead, we find that the continuous shutter of the LiDAR sensor needs to be modeled. Recent radiance field methods (Neurad~\cite{tonderski2024neurad}) accomplish this by modeling each individual LiDAR return. 
Instead, we take inspiration from continuous SLAM literature, which {\em deskews} LidAR scans to componsate for ego motion~\cite{droeschel2018efficient,dellenbach2022ct,park2018elastic,lv2021clins,le2023continuous}. We extend this idea and show how deskewing can 
{\em also} compensate for dynamic actor motion. 

\begin{figure}
    \centering
    \frame{\includegraphics[width=0.22\textwidth,clip,trim= 0 4cm 0 7.05cm]{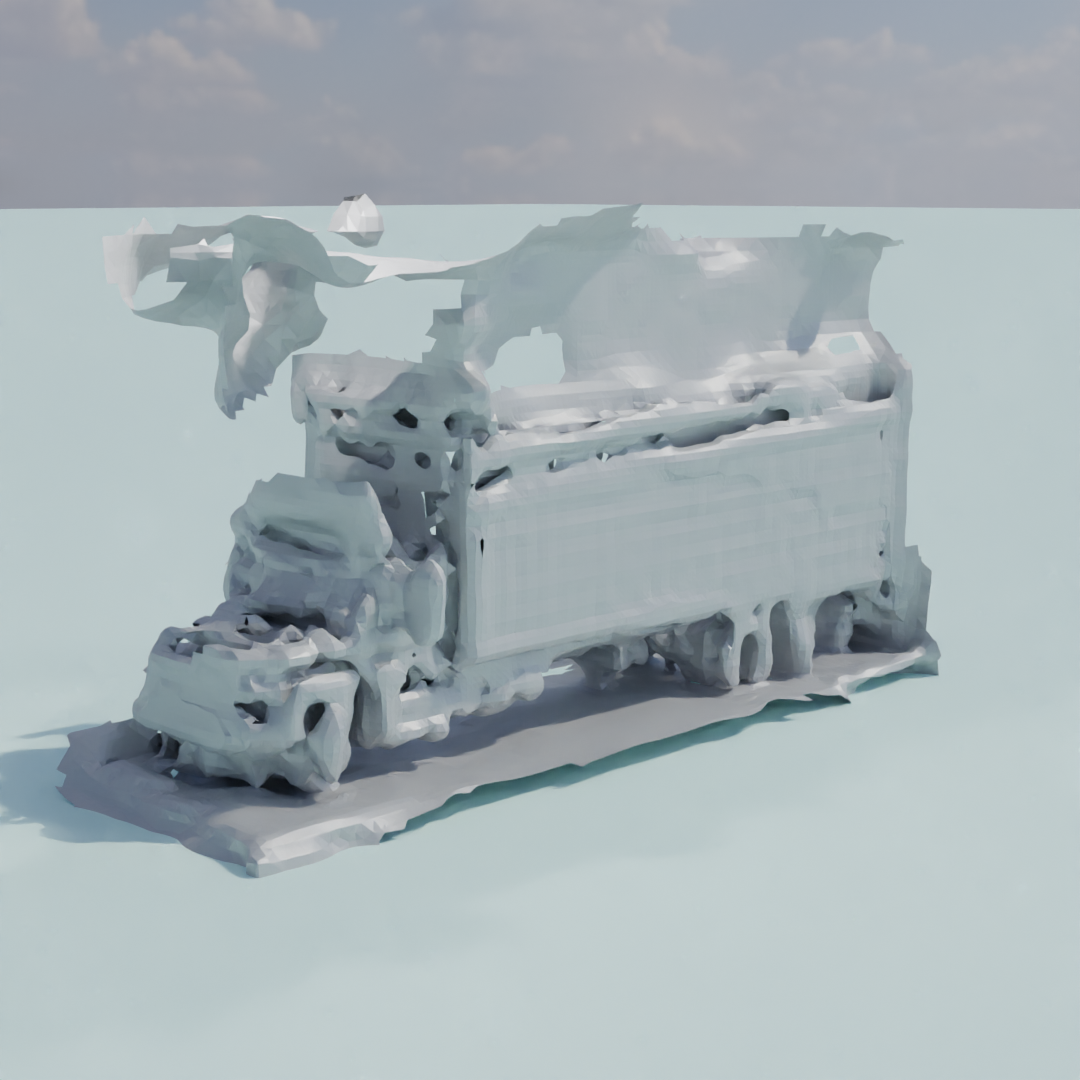}}
    \frame{\includegraphics[width=0.22\textwidth,clip,trim= 0 4cm 0 7.05cm]{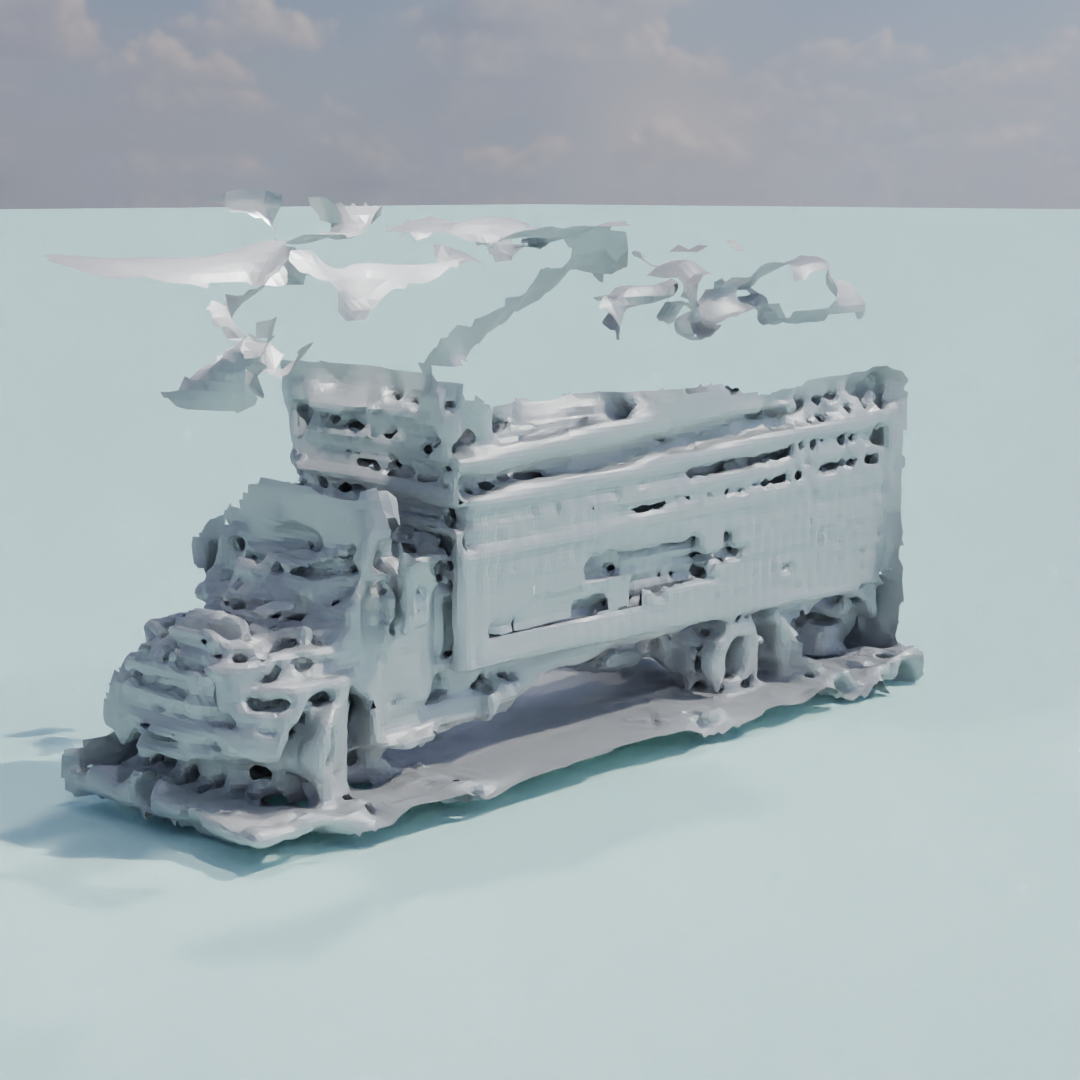}}\\ \vspace{0.1cm}
    \frame{\includegraphics[width=0.45\textwidth,clip,trim= 0 6cm 0 11cm]{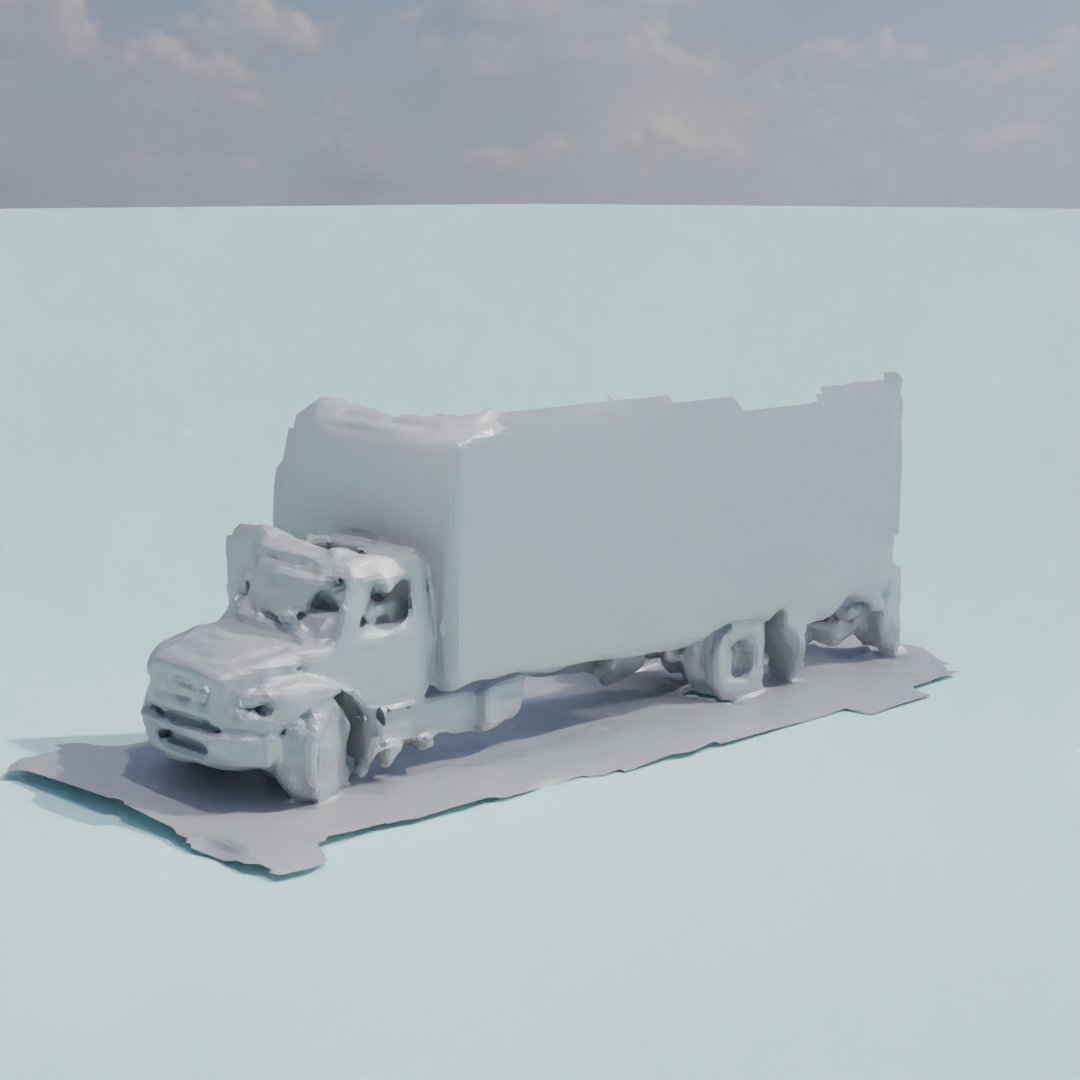}}
    \caption{Dynamic object reconstructions using human-annotated bounding-box annotations ({\bf top left}) tend to be noisy. Optimizing over object pose ({\bf top right}) improves accuracy, while {\em de-skewing} scans to account for dynamic object motion is even more helpful ({\bf bottom}). Please see the animation in the supplement.}
 
 \label{fig:ablation}
\end{figure}
{\bf Applications:} We view this work as a step towards generating dynamic scene reconstructions that provide high-quality annotations for downstream autonomous driving tasks. 
Labeling in-the-wild data is extremely costly, and as a result, many autonomous driving tasks rely on re-processing existing data of varying quality. For example, depth completion benchmarks use aggregated LiDAR sweeps to generate ground truth ``dense" depth reconstructions~\cite{uhrig2017sparsity}. This results in annotated data with well-documented occlusion errors and motion artifacts that are nonetheless still used for training and evaluation~\cite{wang2022cu,zhao2021surface}. An example of the depth maps produced by our method is shown in \cref{fig:depth}
. 
Scene flow is another autonomous driving task that re-processes existing AV datasets, using annotated bounding box motion between frames as a proxy for the underlying ground-truth motion field~\cite{baur2021slim, li2021neural}, which also has well-documented issues in evaluation~\cite{chodosh2024}. Recent scene flow work~\cite{vedder2024zeroflow} has also shown promise when scaled to the large amount of unlabeled LiDAR data that is available~\cite{Argoverse2}. Accurate time-space reconstructions of the rigidly moving objects in the scene are critical to these tasks. We demonstrate that the ground-truth motion annotations are insufficient for producing these reconstructions and that our system significantly improves upon them.


Our main contributions are (1) posing the classic dynamic surface reconstruction problem in the context of LiDAR-based urban scenes, (2) combining insights from actor decomposition of radiance fields and continuous-time SLAM to produce high-quality reconstructions that reduce error by 10X over prior art, and (3) proposing new downstream applications of such spacetime reconstructions.

\section{Related Work}
\label{sec:related}

\textbf{Dynamic Surface Reconstruction:} Reconstruction of non-rigid surfaces from depth scanners has been studied for over two decades\cite{mitra2007dynamic}. Early work overcame the inherent ill-posedness of the problem by relying on object-specific shape models for humans\cite{taylor2012vitruvian}, faces\cite{li2013realtime} and hands\cite{qian2014realtime}. Since then, many works have demonstrated template-free reconstruction in both the online\cite{newcombe2015dynamicfusion} and offline settings\cite{palafox2021npms}. This line
 of work is focused on highly deformable objects such as people and animals, which are very close to the depth sensor. As a result, they do not apply to the long-range, generally rigid world of autonomous driving scenes.

\textbf{Dynamic SLAM:} Since we are solving for the global map of the world as well as the sensor's location within it, our work is closely related to SLAM in general and specifically to Dynamic SLAM, sometimes called SLOT (Simultaneous Localization and Object Tracking). Many works identify dynamic objects to remove them from the global map\cite{hahnel2002map}, but some track dynamic objects and register new observations to an object template. Similar to our work, these approaches typically represent the world as a composition of rigid bodies\cite{bibby2007simultaneous, wang2007simultaneous, dewan2016motion, yang2019cubeslam, henein2020dynamic, tian2023object}. These methods are focused on real-time operation from RGB inputs rather than offline LiDAR processing. As a result, they generally do not reconstruct detailed surface representations of the tracked objects, although that has been proposed as a post-processing step to the tracked objects\cite{jiang2017dynamic}. Also similar to our work are SLAM methods, which create a dense surface reconstruction of the global map\cite{vizzo2021poisson, li2013realtime}. However, to our knowledge, none of these approaches reconstruct dynamic objects. Many works on continuous time SLAM with LiDAR sensors have tackled the rolling shutter problem\cite{dellenbach2022ct,droeschel2018efficient,lv2021clins,le2023continuous}. We take inspiration from their method of deskewing sweeps for ego-motion and apply it to dynamic actors (for the first time, to our knowledge).

\textbf{Asset Generation for Autonomous Driving:} Related to the object reconstruction component of our system is the line of work focused on creating high-quality mesh reconstructions of vehicles for simulation purposes\cite{yang2023reconstructing, manivasagam2020lidarsim, wang2023cadsim}. These methods are similar to ours in that they reconstruct dense meshes of in-the-wild vehicles but have several key differences. First, since these systems aim to extract assets, not reconstruct complete sequences, they focus on objects that are close to the sensor and have accurate poses from object detection. Although this is not made explicit, the result is that these systems are made to operate on stationary objects, not dynamic ones. Second, they rely heavily on RGB information as well as depth sensors. As we show, reconstructing moving objects from spinning LiDARs requires careful handling of the rolling-shutter effect. This makes it challenging to incorporate global-shutter RGB cameras. The fact that these works make no mention of this further indicates that they do not handle dynamic objects.

\textbf{NeRFs} Neural Radiance fields have also been applied to automotive data recently~\cite{ziyang2023snerf,ost021nsg,tonderski2024neurad,turki2023suds}. We take inspiration from the methods which model the world as a composition of rigid actors~\cite{ziyang2023snerf,tonderski2024neurad}. These methods typically also make use of sparse LiDAR signals for geometry estimation, but we find that using an explicit surface-based representation yields superior results.

\section{Problem Statement}
\label{sec:problem}
We assume as input a sequence of LiDAR sweeps measured at timestamps $t \in \mathcal{T}$, and coarse tracks of $K$ objects. Since we are using a compositional model of the scene, we will need a coordinate frame for each component.
\begin{itemize}
    \item \textbf{Ego coordinates:} This is the moving ego-vehicle coordinate frame used to measure input points, 
    denoted as $e_t$.
    \item \textbf{Object coordinates:} 
    Each dynamic object $i$ is assigned its own canonical coordinate frame at time $t$ where the $x$-direction is ``forward", denoted as $o_t^i$.
    \item \textbf{World coordinates:} We represent the static background in a fixed world coordinate frame $w$, which we set equal to $e_1$ without loss of generality. 
\end{itemize}

We use subscripts to denote the coordinate frame of given point $\mathbf{x}$ or set of points $\mathbf{X}$: e.g., $\mathbf{x}_{e_t}$,$\mathbf{X}_{e_t}$. 
We express the rigid transformation between coordinate frames as $\mathbf{T} \in R^{4 \times 4}$; 
e.g., the transformation from world to sensor coordinates $e_t$ is $\mathbf{T}^{e_t}_w$, while the transformation from object to world coordinates is $\mathbf{T}_{o^{i}_t}^w$. Then, transformation from object to sensor coordinates at time $t$ is $\mathbf{T}^{e_t}_{o^i_t} = \mathbf{T}^{e_t}_w \mathbf{T}_{o^{i}_t}^w$. 

We aim to decompose the scene into a set of surfaces that transform rigidly over time. Our approach is agnostic to the particular choice of surface representation, but we use triangular meshes since they are lightweight and widely used. We write the surface (mesh) of each of $K$ objects in the scene $\{\mathcal{M}_i\}_{i=1}^{K}$ as well as the background ``object" $\mathcal{M}_0$. In a slight abuse of notation, we write $\mathbf{T}\mathcal{M}$ to denote the surface $\mathcal{M}$ transformed by $\mathbf{T}$ and write $\mathbf{T}\mathbf{X}$ to express the transformed points $\mathbf{X}$. We write the composition of 2 surfaces as $\left[ \mathcal{M}_1, \mathcal{M}_2 \right]$. Finally, we will measure the 3D distance between a surface and a point cloud using the nearest neighbor loss
\begin{equation}
    \mathcal{D}(\mathcal{M}, \mathbf{X}) = \sum\limits_{\mathbf{x} \in \mathbf{X}} \min\limits_{\mathbf{m} \in \mathcal{M}} \lVert \mathbf{m} - \mathbf{x} \rVert . \label{eq:nn}
\end{equation}


\section{Objective}
Our method aims to find the surfaces and object motions that best explain the LiDAR measurements. Using the notation from the previous section, we can express this goal as the optimization objective:
\begin{equation}
\label{eq:global}
    \min_{\{\mathcal{M}_i, \mathbf{T}_{o^i_t}^{e_t}, \mathbf{T}_w^{e_t}\}} \sum\limits_{t \in \mathcal{T}} \mathcal{D}(\left[\mathbf{T}_w^{e_t} \mathcal{M}_0, \mathbf{T}_{o^1_t}^{e_t} \mathcal{M}_1, \ldots,\mathbf{T}_{o^K_t}^{e_t} \mathcal{M}_K \right], \mathbf{X}_{e_t}).
\end{equation}
Intuitively, this equation sums up, for all $t$, the error between the point cloud at time $t$ and the reconstruction at time $t$. Each term in the summation uses the object and ego motions ($\mathbf{T}$) to compose all of the meshes ($\mathcal{M}$) into a reconstruction at time $t$. The reconstruction is then compared to the point cloud at that time ($\mathbf{X}_{e_t}$) using the error metric $\mathcal{D}$.

\subsection{Decomposition}
\label{sec:decomp}
Our approach consists of applying coordinate descent to \Cref{eq:global}: alternating between fixing the poses to {\bf optimize surfaces} and then fixing the surfaces to {\bf optimize poses}. This approach allows us to leverage off-the-shelf tools for each step of the optimization. 
In the following sections we derive the appropriate surface and pose optimization steps from the global objective.

The first step of both derivations decomposes the objective across objects. Let $\mathbf{X}^{i}_{e_t}$ denote the subset of points from $\mathbf{X}_{e_t}$ which fall on object $i$. This assignment can be coarse, and in practice, we assign points to the bounding box they fall into and to the background if they are not contained in any bounding box. Using this notation, we can further break down \cref{eq:global} into:
\begin{equation}
\label{eq:global-assigned}
    \min_{\{\mathcal{M}_i, \mathbf{T}_{o^i_t}^{e_t}, \mathbf{T}_w^{e_t}\}} \sum\limits_{t \in \mathcal{T}}\sum\limits_{i=0}^K \mathcal{D}(\mathbf{T}_{o^i_t}^{e_t} \mathcal{M}_i, \mathbf{X}^i_{e_t}),
\end{equation}
where we let $o^{0}_t = w$ for notional simplicity. We can now derive the pose and surface optimization steps.
\subsection{Optimize surfaces}

Assuming fixed poses (initialized from LiDAR odometry and object tracks), we estimate new surfaces by solving
\begin{equation}
\label{eq:mesh-step0}
    \mathcal{M}_i \mapsfrom \arg\min\limits_{\mathcal{M}_i} \sum\limits_{t \in \mathcal{T}}\mathcal{D}(\mathbf{T}_{o^i_t}^{e_t}\mathcal{M}_i, \mathbf{X}^i_{e_t}).
\end{equation}

We make use of two identities to simplify this optimization. First, we use the fact the distance is unaffected by a global rigid transformation to see that $\mathcal{D}(\mathbf{T}\mathcal{M}, \mathbf{X}) = \mathcal{D}(\mathcal{M}, \mathbf{T}^{-1}\mathbf{X})$. Second, if we write a set of points $\mathbf{X} = [\mathbf{X}_1, \mathbf{X}_2]$ as a union of two disjoint sets $\mathbf{X}_1$ and $\mathbf{X}_2$, we can see that $\mathcal{D}(\mathcal{M}, [\mathbf{X}_1, \mathbf{X}_2]) = \mathcal{D}(\mathcal{M}, \mathbf{X}_1) + \mathcal{D}(\mathcal{M}, \mathbf{X}_2)$. Now we combine them to get:
\begin{equation}
\label{eq:mesh-step-alg}
\begin{split}
     \mathcal{M}_i &\mapsfrom \arg\min\limits_{\mathcal{M}_i} \sum\limits_{t \in \mathcal{T}}\mathcal{D}(\mathbf{T}_{o_t^i}^{e_t} \mathcal{M}_i, \mathbf{X}^i_{e_t})\\
     &= \arg\min\limits_{\mathcal{M}_i} \sum\limits_{t \in \mathcal{T}}\mathcal{D}(\mathcal{M}_i, (\mathbf{T}_{o_t^i}^{e_t})^{-1}\mathbf{X}^i_{e_t})\\
     &= \arg\min\limits_{\mathcal{M}_i} \mathcal{D}\left(\mathcal{M}_i, \left[(\mathbf{T}_{o_t^i}^{e_t})^{-1}\mathbf{X}^i_{e_t}, \ldots\right]\right).\\
\end{split}
\end{equation}

The final form of this equation can be interpreted as a standard static point-to-surface reconstruction problem. We use the recent Neural Kernel Surface Reconstruction~\cite{huang2023nksr}, but any technique, such as Poisson surface reconstruction~\cite{kazhdan2006poisson}, could be used.

\subsection{Optimize poses}
Assuming fixed surfaces, we can estimate new poses by solving
\begin{equation}
\label{eq:pose-step0}
\begin{split}
    \mathbf{T}^{e_t}_{o_t^i} &\mapsfrom \arg\min\limits_{\mathbf{T}^{e_t}_{o_t^i}} \mathcal{D}\left(\mathbf{T}_{o^i_t}^{e_t}\mathcal{M}_i, \mathbf{X}^i_{e_t}\right)\\
    &= \arg\min\limits_{\mathbf{T}^{e_t}_{o_t^i}} \mathcal{D}\left(\mathcal{M}_i, (\mathbf{T}_{o^i_t}^{e_t})^{-1}\mathbf{X}^i_{e_t}\right).\\
\end{split}
\end{equation}
This is a point-to-surface registration problem that is well-studied under the family of Iterative Closest Point (ICP) methods. However, there is an remaining complication due to the rolling shutter of LiDAR capture. 
\begin{figure}
    \centering
    \frame{\includegraphics[width=0.45\textwidth,clip, trim = 0cm 6cm 0cm 10cm]{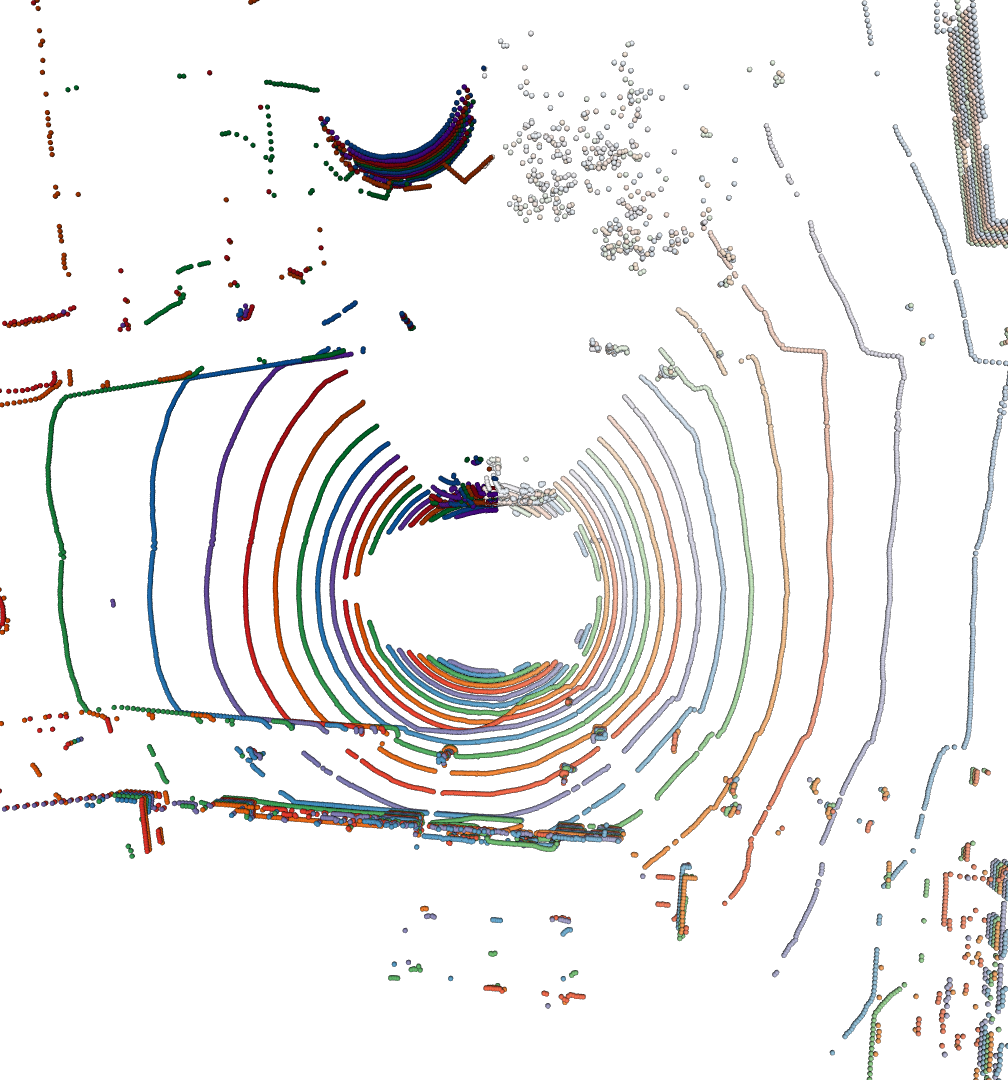}}\hspace{0.05cm}
    \frame{\includegraphics[width=0.45\textwidth, clip, trim = 0 3cm 0cm 5cm]{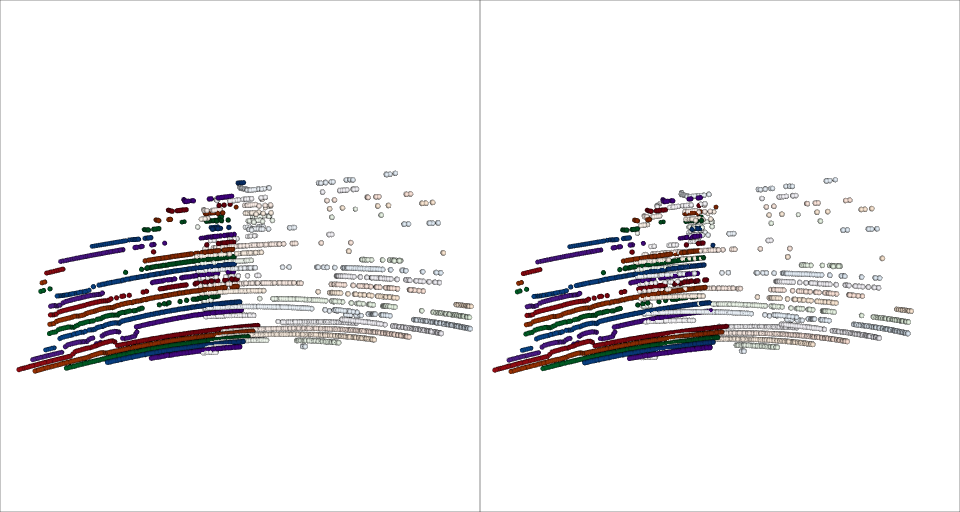}}
    \caption{(\textbf{Left}) A LiDAR sweep where each point has been colored according to which laser it belongs to (hue) and the time within the sweep it was acquired (lighter is earlier, darker is later). A moving car is passing the ego-vehicle (traveling right) on the left and is captured at both the start and end of the sweep (\textbf{top right}), leading to distortion (the driver-side window is captured twice in different locations). Accounting for this distortion by modeling the object motion is key to the quality of our reconstructions (\textbf{bottom right}).}
    \label{fig:what-is-a-sweep}
\end{figure}
\subsection{What {\em is} a LiDAR sweep?}
\begin{figure*}[t]
    \centering
    \frame{\includegraphics[width=0.31\textwidth]{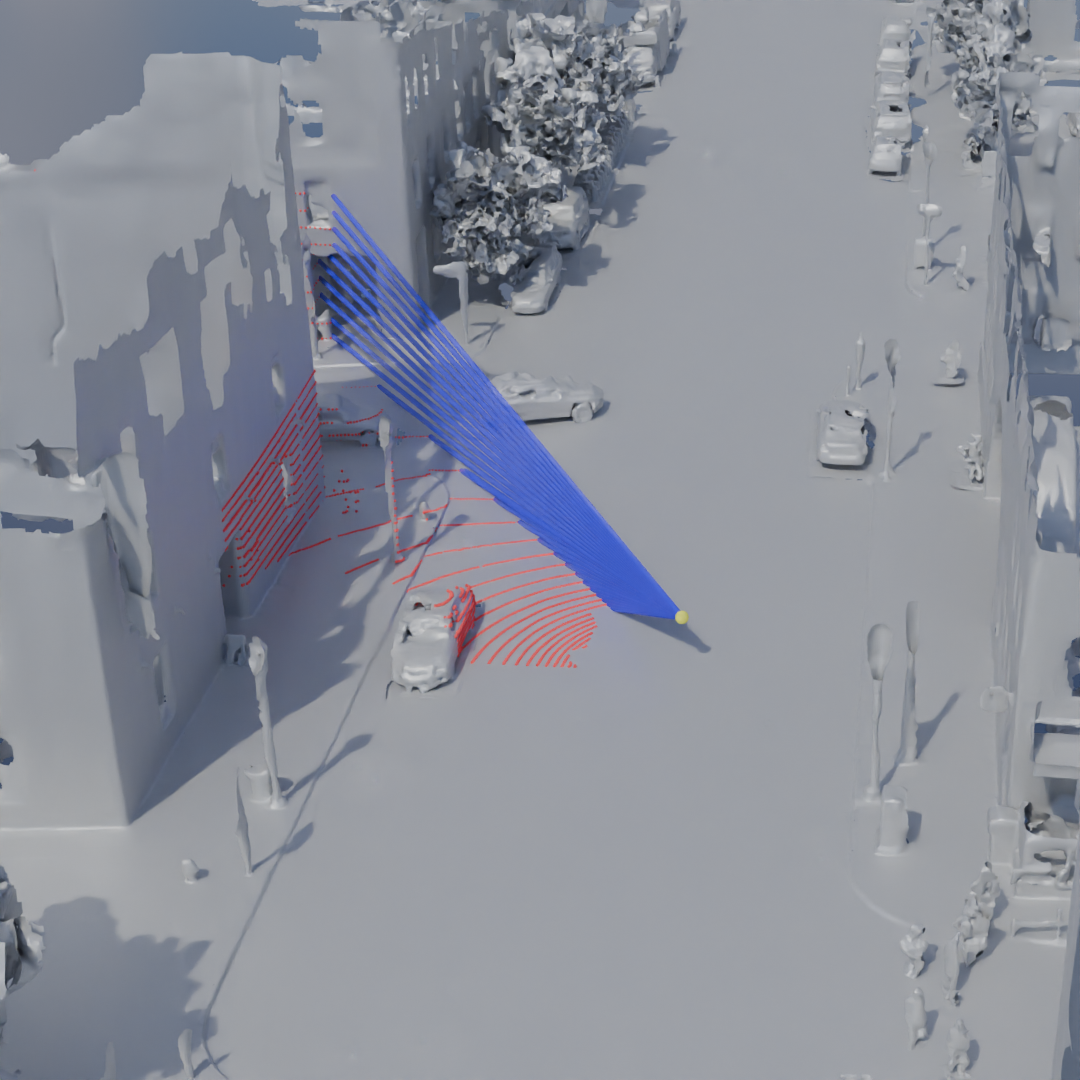}}\hspace{0.1cm}
    \frame{\includegraphics[width=0.31\textwidth]{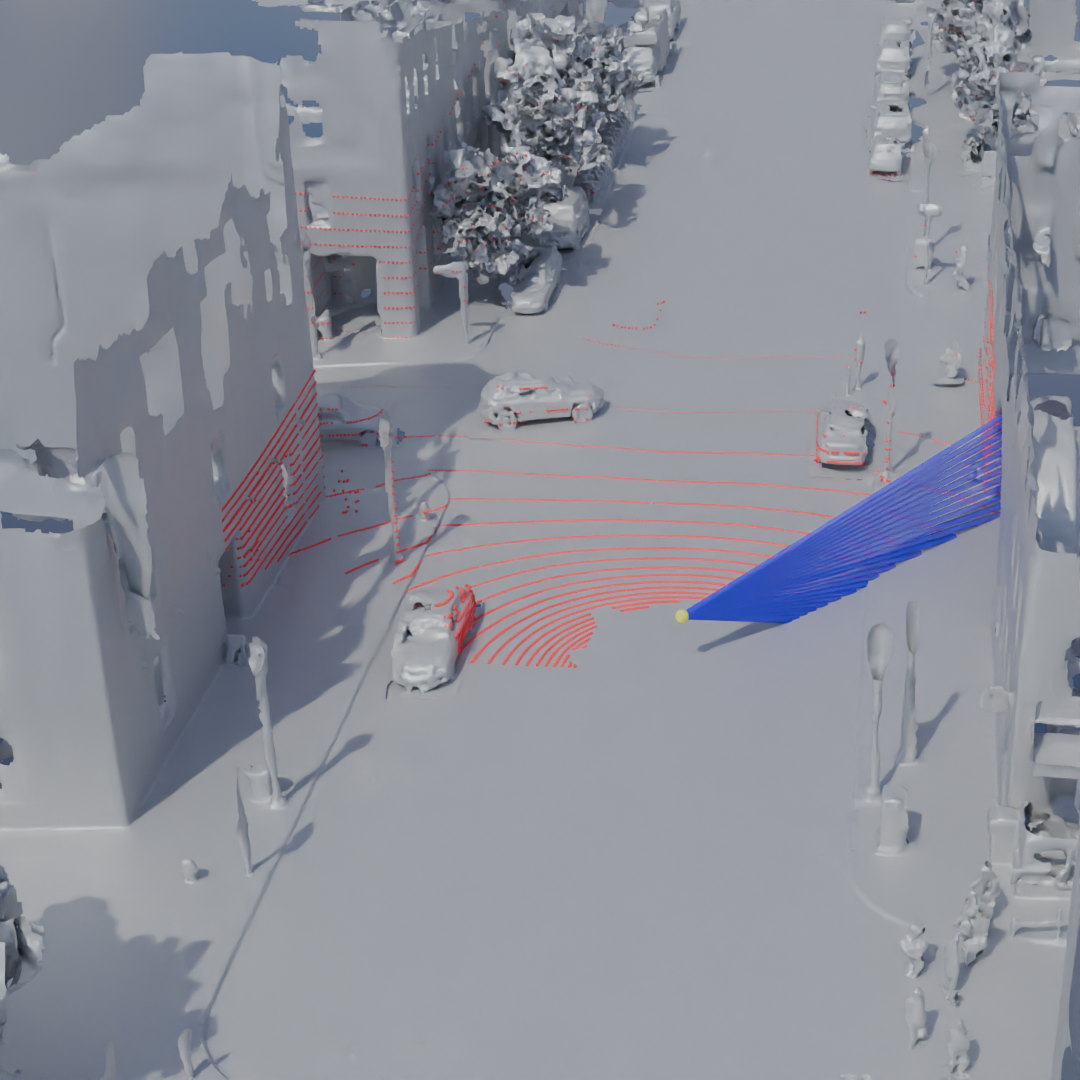}}\hspace{0.1cm}
    \frame{\includegraphics[width=0.31\textwidth]{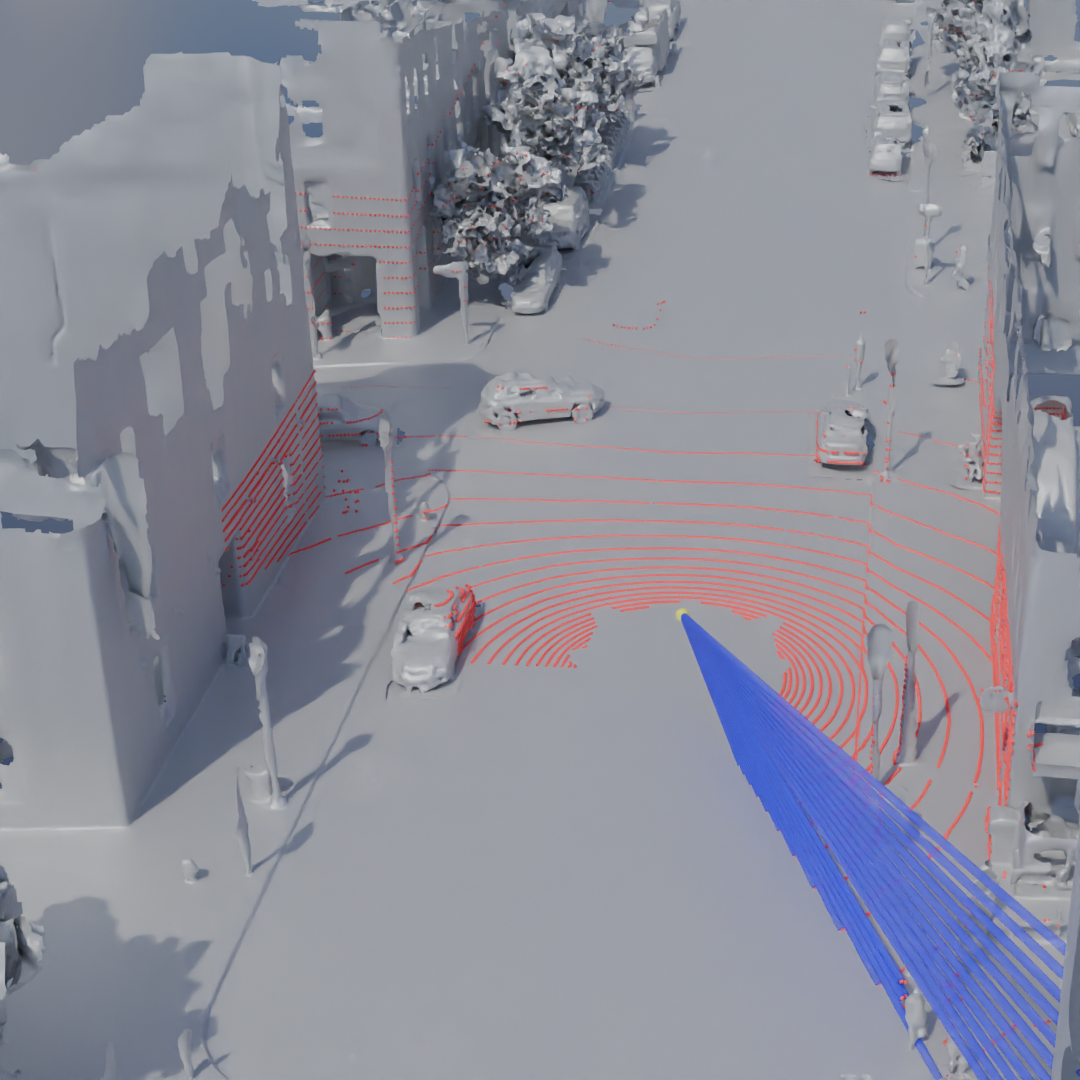}}\hspace{0.1cm}\\\vspace{0.1cm}
    \frame{\includegraphics[width=0.31\textwidth]{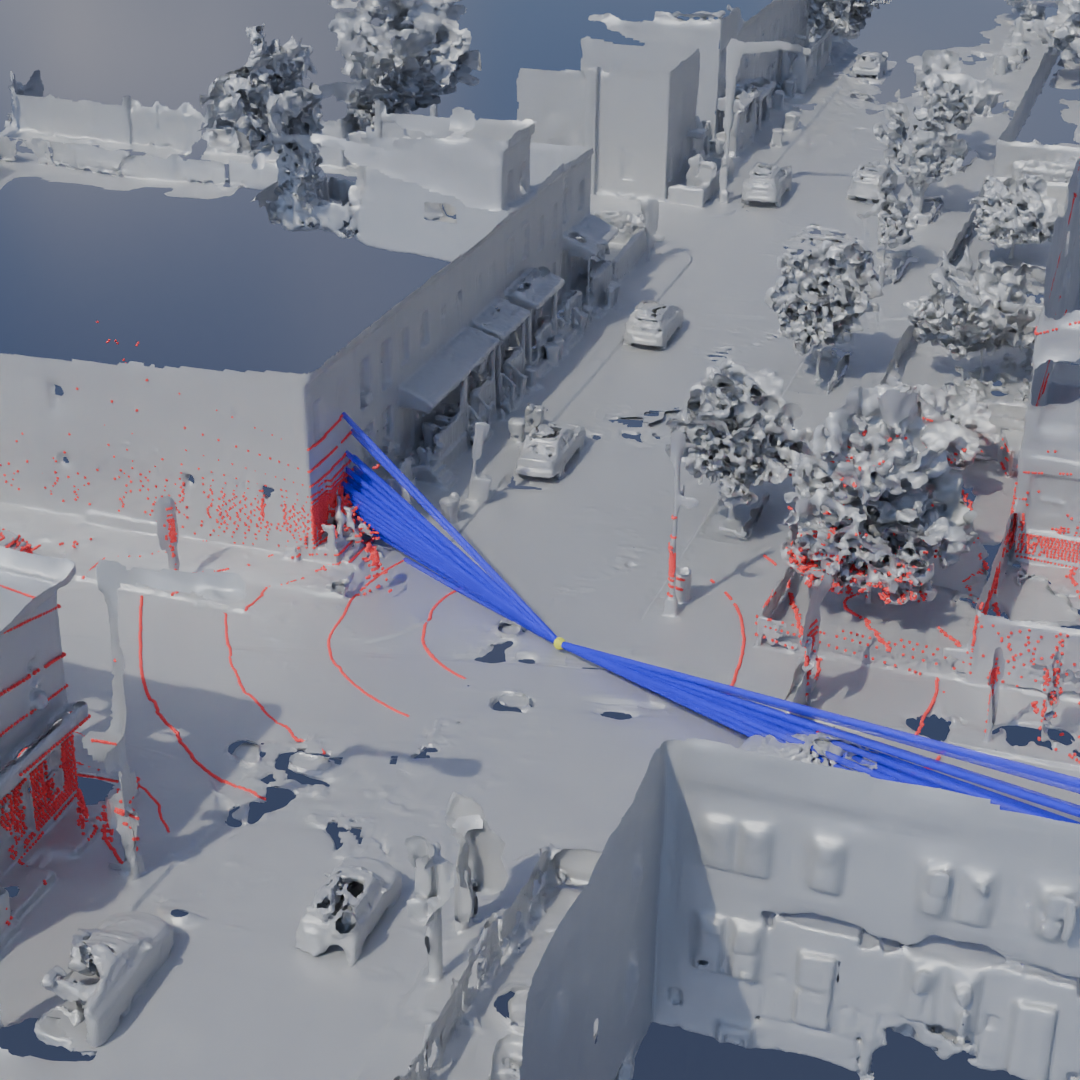}}\hspace{0.1cm}
    \frame{\includegraphics[width=0.31\textwidth]{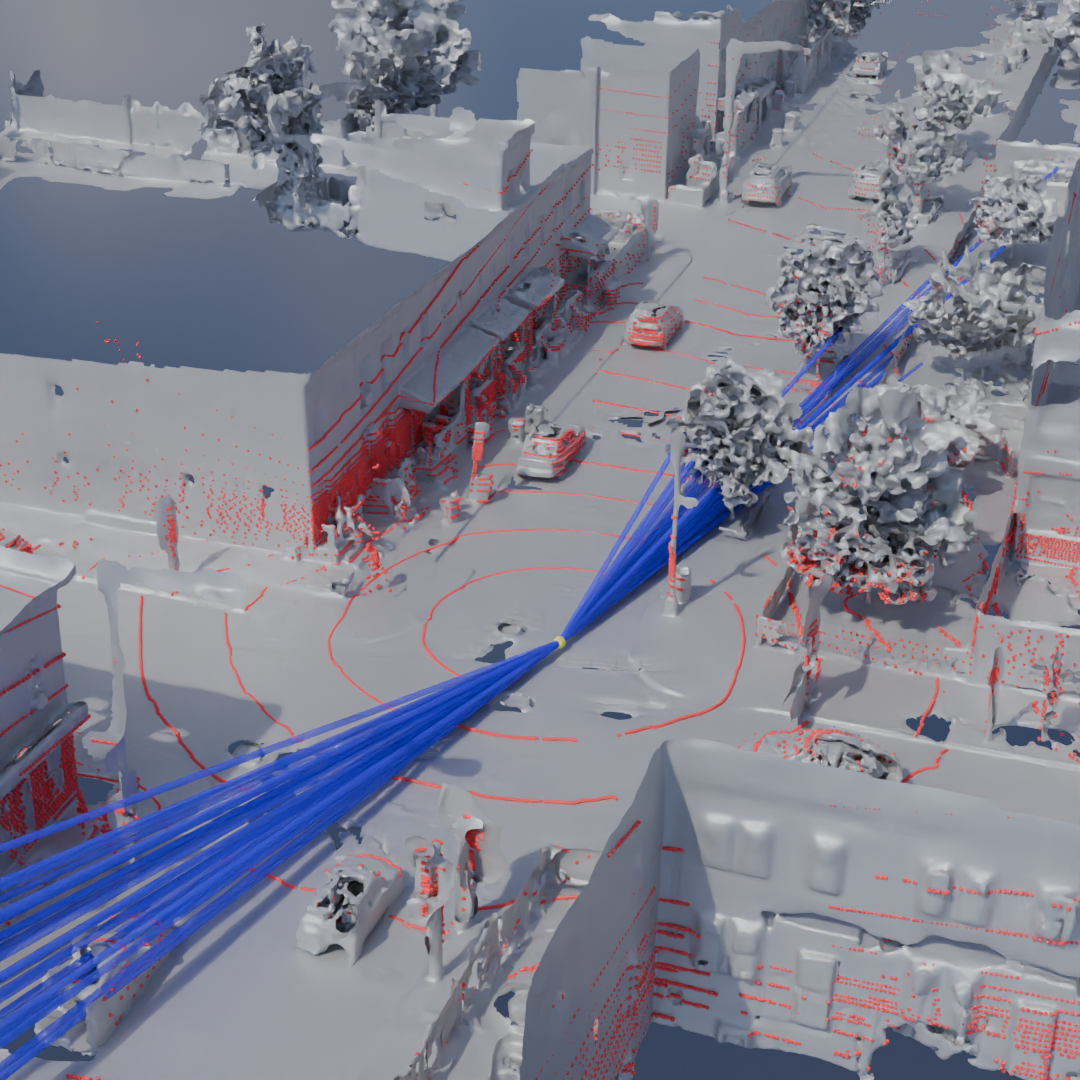}}\hspace{0.1cm}
    \frame{\includegraphics[width=0.31\textwidth]{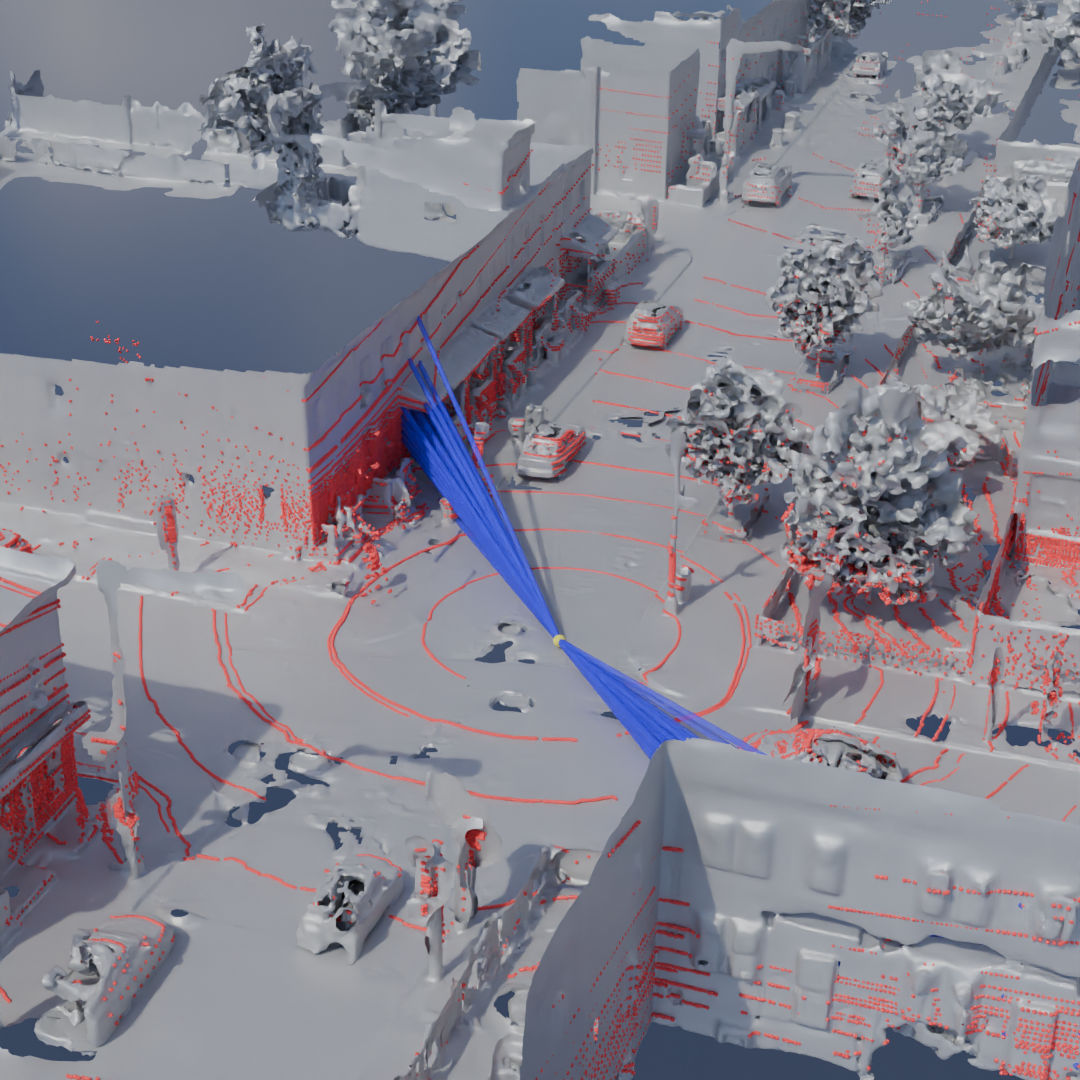}}\hspace{0.1cm}\\
    \caption{LiDAR is often abstracted as 360-degree sweeps captured with a global shutter, but is actually captured with a continuous rotating shutter from a moving ego-car. Our continuous-time optimization framework  correctly models this, 
    dramatically improving the quality of urban reconstructions. Here, we visualize the set of rays captured at a time instant (blue lines) for NuScenes (\textbf{top}) and Argoverse~\cite{Argoverse2} (\textbf{bottom}). Interestingly, our approach is even more effective for recent AV datasets~\cite{Argoverse2,sun2020scalability} that employ {\em multiple} spinning lidars, which are often set to be out-of-phase to minimize interference (but adding to the inconsistency of a global shutter approximation).}
    \label{fig:ray-vis}
\end{figure*}
Revolving LiDAR sensors do not have a global shutter. Instead, they rotate continuously and measure depth across 16-128 vertically arranged lasers, typically taking 100ms to complete a 360-degree rotation. Depth along each laser ray is measured with respect to the (potentially moving) ego sensor frame (see \cref{fig:ray-vis}). Most datasets abstract away this continuous capture and instead provide LiDAR returns as groups of 360-degree sweeps. This is convenient for many downstream applications but introduces a few subtle issues which we hope to shed some light on.

\textbf{Ego-Motion Distortion:} When the sensor moves during a sweep the resulting point cloud is distorted. This distortion has been removed in all of the popular LiDAR datasets through ``deskewing'', which transforms all of the measured points into the sensor's reference frame at the beginning or end of the sweep. We can use our existing notation to explain this process by letting our time index $t$ be a continuous variable rather than an integer frame index. For example, if we have a 16-beam LiDAR sensor that takes 1080 measurements in a single rotation, the first 16 points of our sequence are written as $\mathbf{X}_{e_{1/1080}}$. Importantly, our global optimization \cref{eq:global}, mesh step \cref{eq:mesh-step0}, and pose step \cref{eq:pose-step0} are just as valid under this interpretation of a sweep ``slice'', but with 1080 times as many poses. This is computationally expensive and may underconstrain the optimization. To avoid this, we adopt a constant velocity model for poses between ``keyframes'' placed at the end of every complete sensor rotation. For example, we can express the continuous pose of the sensor for $0 < t < 1$ using the keyframe poses $\mathbf{T}_{e_0}^{w}, \mathbf{T}_{e_1}^{w}$ like so:
\begin{equation}
\label{eq:ego-deskew}
\begin{gathered}
\begin{aligned}
    \mathbf{T}_{w}^{e_1} \left(\mathbf{T}_w^{e_0}\right)^{-1} = \begin{bmatrix}
        \mathbf{R}_{3 \times 3} & \mathbf{v}_{3 \times 1}\\
        \mathbf{0}_{1 \times 3} & 1
    \end{bmatrix},&\quad
    \mathbf{w} = \log\left(\mathbf{R}\right)\\
\end{aligned}\\
    \mathbf{T}_{e_t}^{w} = \mathbf{T}_{e_1}^w\begin{bmatrix}
        e^{\mathbf{w}(1 - t)} & \mathbf{v}(1 - t)\\
        \mathbf{0}_{1 \times 3} & 1
    \end{bmatrix} = \mathbf{T}_{e_1}^{w} \mathbf{T}_{e_t}^{e^1}.
\end{gathered}
\end{equation}
Using these equations, points can be deskewed by transforming all of the points in a sweep to the coordinate frame $e_0$. {\em While deskewing generates the correct virtual point cloud for a static world, it fails to compensate for moving objects and complicates reasoning about the free space}. 

\textbf{Actor-Motion Distortion:} Unlike ego-motion distortion, actor-motion distortion is still present in essentially all AV datasets. However, our spacetime optimization can correctly model moving objects by applying the same insight; 
just as we assumed that the ego-vehicle obeys a constant velocity model between keyframes, we can make the same assumption about {\em other} moving objects. However, care needs to be taken to ensure that the constant velocity assumption is applied to the object's motion \emph{with respect to the world} as opposed to with respect to the ego-vehicle. Constant velocity in the world frame is not equivalent to constant velocity in the moving sensor frame due to the presence of rotations. With this in mind, we represent the object poses like so:
\begin{equation}
\label{eq:obj-deskew}
\begin{gathered}
\begin{aligned}
    \left(\mathbf{T}_{e_1}^{o_1}\mathbf{T}^{e_1}_{w}\right) \left(\mathbf{T}_{e_0}^{o_0}\mathbf{T}^{e_0}_{w}\right)^{-1} = \begin{bmatrix}
        \mathbf{R}_{3 \times 3} & \mathbf{v}_{3 \times 1}\\
        \mathbf{0}_{1 \times 3} & 1
    \end{bmatrix}
    &,\quad\mathbf{w} = \log\left(\mathbf{R}\right)
\end{aligned}\\
    \mathbf{T}_{e_t}^{o^i_{t}} = \begin{bmatrix}
        e^{\mathbf{w}(1 - t)} & \mathbf{v}(1 - t)\\
        \mathbf{0}_{1 \times 3} & 1
    \end{bmatrix}\mathbf{T}_{e_1}^{o_1^i}\mathbf{T}_{e_t}^{e_1} = \mathbf{T}^{o^i_t}_{o^i_1}\mathbf{T}_{e_1}^{o_1^i}\mathbf{T}_{e_t}^{e_1}.
\end{gathered}
\end{equation}
This factorization is not only the correct way of applying the constant velocity assumption but also makes it easy to deal with the fact that, in many cases, public datasets only release the ego-motion compensated points $\mathbf{T}_{e_t}^{e_1}\mathbf{X}_{e_t}$. With the above factorization, we can omit the first $\mathbf{T}_{e_t}^{e_1}$ transformation as it has already been applied. This is one of those fortunate situations where the easy and correct approaches are the same! In our algorithm, both ego and actor distortion can be accounted for by plugging \cref{eq:ego-deskew} and \cref{eq:obj-deskew} into our mesh and pose steps.

\textbf{Freespace errors:} Finally, we point out that the naive deskewing common in existing datasets may introduce a subtle ``error'': the deskewed point cloud no longer reveals the true free space in the scene. Many radiance field~\cite{deng2022depth,manivasagam2020lidarsim,huang2023neural} and surface reconstruction~\cite{huang2023nksr} methods explicitly reason about the free space travelled by a LiDAR pulse along a ray from the sensor to a surface.
However, such constraints no longer hold after deskewing. Our approach handles this problem by explicitly reasoning about the correct (time-varying) origin for each ray, and therefore the implied freespace, for each point. We provide additional visuals in the supplement. 





\section{Experiments}
\textbf{Datasets:} All of our experiments are conducted on nuScenes\cite{caesar2020nuscenes} and Argoverse 2.0\cite{Argoverse2}. We focus primarily on nuScenes as its noisy annotations and sparse LiDAR present the greatest challenge to accurate geometry recovery. The nuScenes experiments are conducted on the logs used in the miniature set released by the dataset authors (we omit scene-0553 since it does not contain any ego-motion), except for the NVS experiment, which was conducted on scene-0103 due to computation constraints. The Argoverse 2.0 experiments are conducted on a similarly sized subset of the validation set; sequences are listed in the supplement.

\subsection{Lidar Novel View Synthesis} 
\begin{figure}
\centering
\frame{\includegraphics[width=0.22\textwidth,clip,trim=0 1cm 0 3cm]{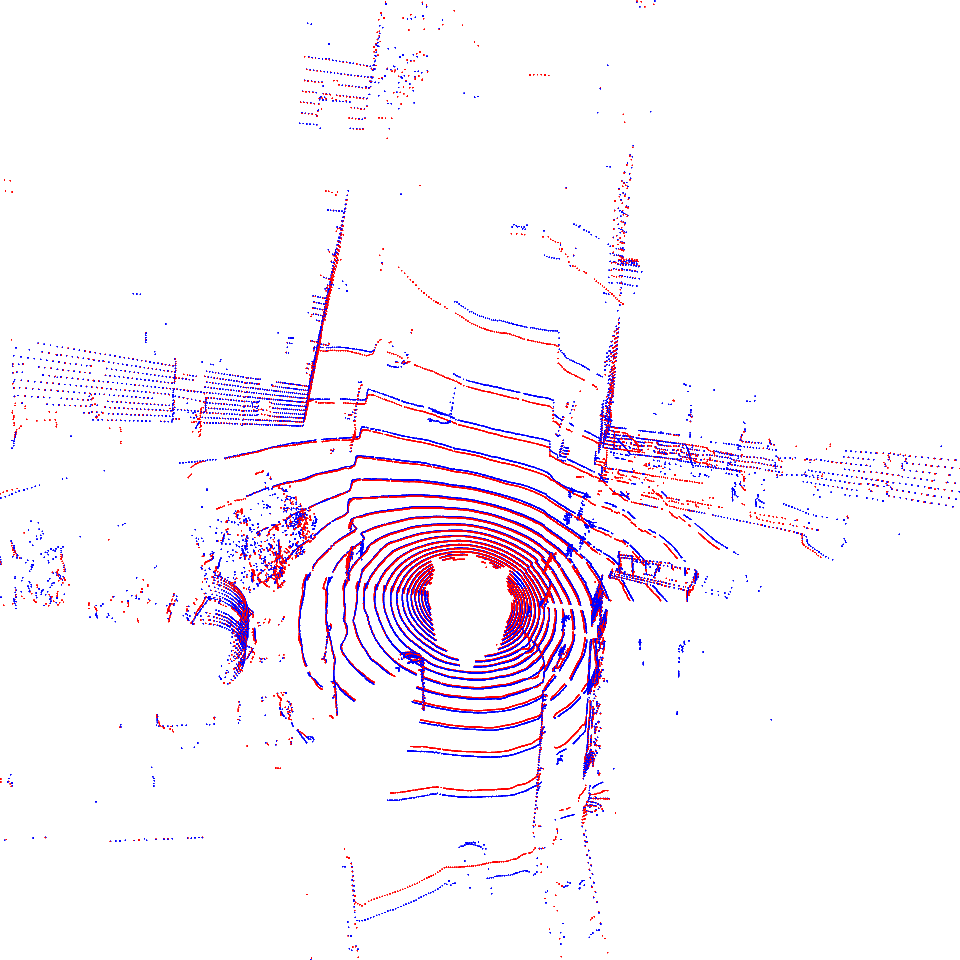}}\vspace{0.1cm}
\frame{\includegraphics[width=0.22\textwidth,clip,trim=0 1cm 0 3cm]{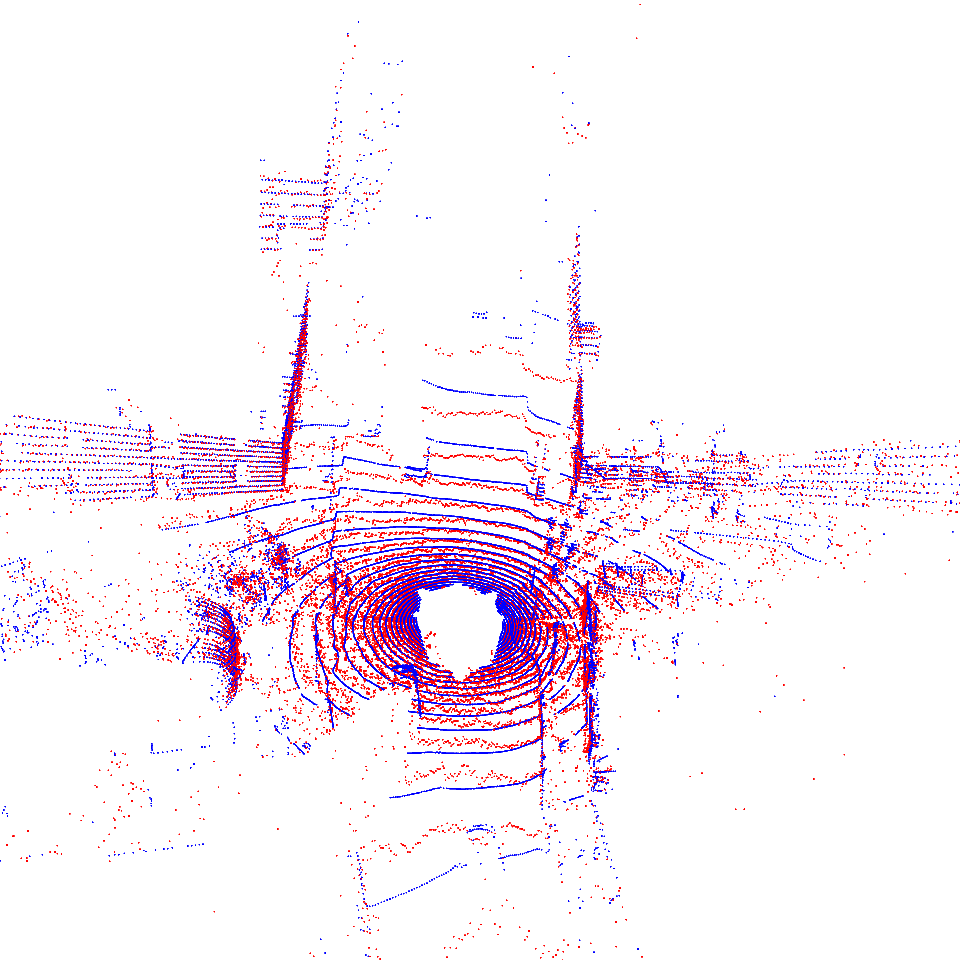}}
\caption{Examples of the synthesized point clouds from our method (\textbf{left}) versus NeuRAD~\cite{tonderski2024neurad} (\textbf{right}). Ground truth is shown in blue and the predicted point clouds are shown in red. We see that the density-based geometry of \cite{tonderski2024neurad}, while producing high-quality RGB renderings, is uneven, leading to high errors on flat oblique surfaces such as the road.}
\label{fig:pcd}
\end{figure}

Following previous work, we use LiDAR Novel View synthesis for our primary evaluation\cite{huang2023neural,tonderski2024neurad}. In this task, we withhold 10\% of the LiDAR sweeps when training our model and then aim to predict the held-out sweeps. We compare our method and NeuRAD\cite{tonderski2024neurad} on this task by evaluating the chamfer distance and median L2 distance error between the synthesized point cloud and ground truth point cloud. Since the ground truth poses for both the ego-vehicle and actors are noisy, we face the same challenge as many NeRF-in-the-wild works\cite{zhang2021ners,lin2021barf}: what pose to use when running test queries? We follow \cite{zhang2021ners} and let each method use the test pose that achieves the lowest error.\\
\textbf{SMORE Details:} We run the SMORE optimization on the training views to obtain reconstructed meshes for objects and the background. Then, we optimize the test poses to by running one pose step (ICP). Finally, we iterate over the test sweeps to synthesize point clouds by computing ray-mesh intersections. We initialize the object tracks and bounding boxes either by using linear interpolation on the provided object annotations or with the output of an off-the-shelf LiDAR object tracker\cite{peri2023towards}. We initialize the ego-pose using an off-the-shelf LiDAR odometry method\cite{vizzo2021poisson}. We then run 100 refinement iterations on all objects and background maps, with early stopping criteria to avoid wasted computation. Iterations are stopped if the mean registration error for an object falls below 1 centimeter for three consecutive iterations. For the mesh step, we use the default parameters of the publicly released Neural Kernel Surface Reconstruction model\cite{huang2023nksr}. For the pose step, once we have deskewed the dynamic objects, we use a standard ICP implementation\cite{Zhou2018} with a point-to-plane loss, a robust Huber kernel with $k=0.2$ and a matching threshold of $1.5$ meters. For the tracking results, we use the Centerpoint-based\cite{yin2021center} object detector LT3D\cite{peri2023towards} to extract bounding boxes in all frames. We then use greedy association to turn the detections into object tracks.\\
\textbf{NeuRAD Details:} We train NeuRAD on the training split using the same initialization as our method. For testing, however, the reference implementation does not support optimizing new poses that were not present at train time. Instaed, for each test sweep, we interpolate between the closest optimized training poses. Alternatively, we completely disable pose optimization and use the final optimized poses from our method. Note that, by default, NeuRAD also optimizes its dynamic actor poses. In this setting, we observed worse results when running train pose optimization. In the results shown, we fix actor poses from the initialization. Finally, we synthesize a point cloud by computing the expected depth along the rays given by each test sweep.\\
\textbf{Results}: In \cref{tab:nvs}, we see that SMORE outperforms NeuRAD by order of magnitude in both chamfer distance and median depth error. This is the case even when NeuRAD is given the final optimized poses from our method, which we note do outperform the poses found through NeuRAD's optimization. We observe that the error persists due to rays going through ``holes'' in the density field, causing outlier points and uneven surface geometry, leading to high errors on oblique surfaces such as the road. See \cref{fig:pcd} for an example of the latter phenomenon.
\subsection{Pose Estimation}
\textbf{Actor Pose:} A key advantage of our method is its ability to recover from large errors in the input actor and ego poses. To test this, we run our method with annotations taken at various sample rates. Then, we compare our method's predicted object locations to the unsampled ground truth annotations using Average Translation Error metric\cite{caesar2020nuscenes}. We compute this metric over the nine test sequences and filter out objects that follow linear trajectories. As shown in \cref{tab:actor-pose}, our method is robust even to extreme subsampling and consistently improves over the linearly interpolated poses, which are often used for downstream tasks\cite{chodosh2024}. We also evaluate the surface quality of our method across the different subsampling rates and when using an off-the-shelf tracker. The results in \cref{tab:surf} further confirm the robustness of our method to input annotation errors.\\
\textbf{Ego Pose:} We wish to evaluate our method's ability to recover accurate ego-poses. However, the ground truth poses are too noisy to make a straightforward ground truth comparison useful. Instead, we show that the poses from our method are superior to the ground truth and \cite{vizzo2023ral} by training our NeRF baseline\cite{tonderski2024neurad} on all three sets of poses. In \cref{tab:ego-pose-neurad}, we can see that our poses result in far better scene reconstruction. We note that the large disparity between the NeuRAD results when trained on the exact same poses as our method (last row) confirms that the improvement in our method is not only due to better poses but also better geometry estimates.
\begin{figure}
    \centering
    \frame{\includegraphics[width=0.22\textwidth,clip,trim = 0cm 7cm 0cm 8cm]{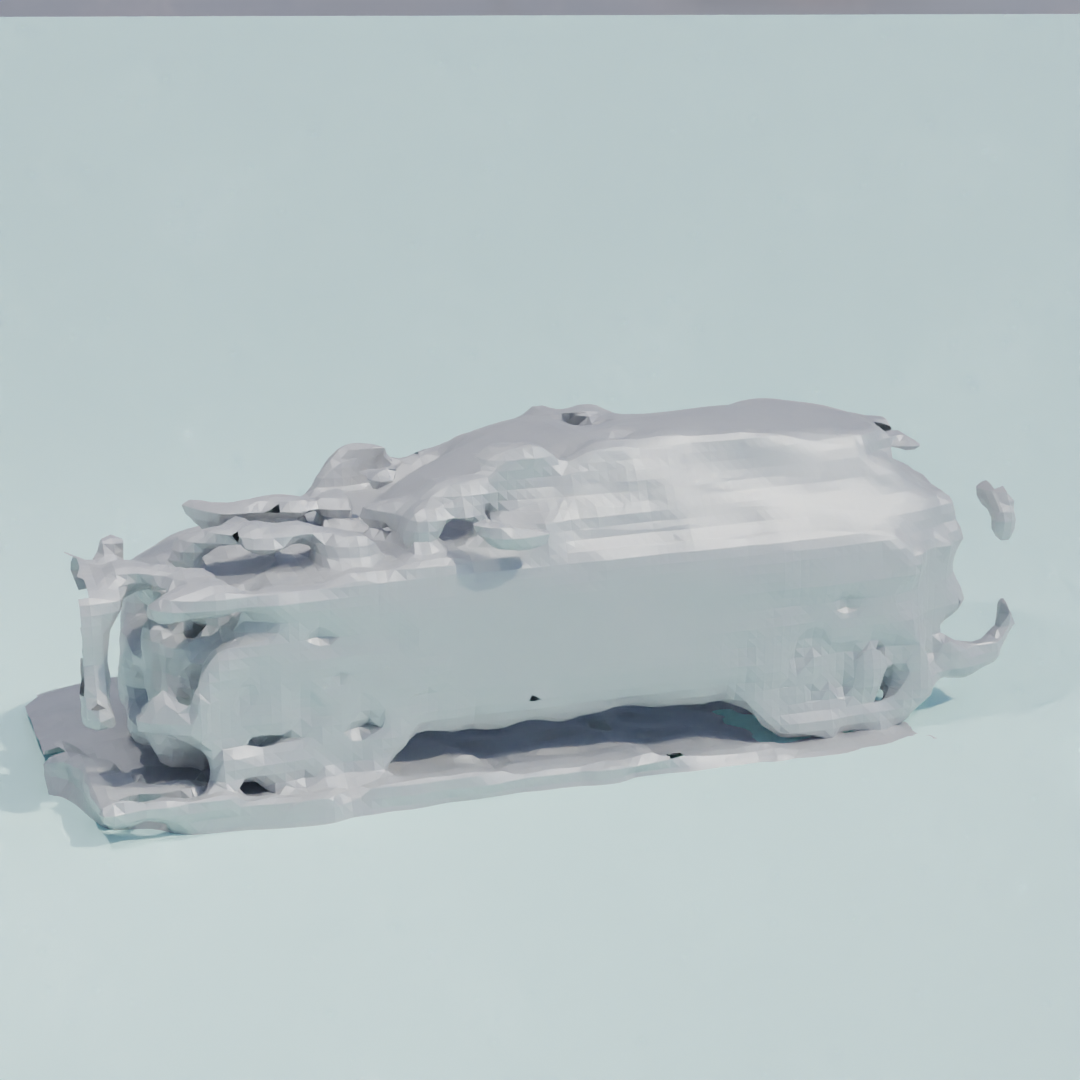}}\hspace{-0.05cm}
    \frame{\includegraphics[width=0.22\textwidth,clip,trim = 0cm 7cm 0cm 8cm]{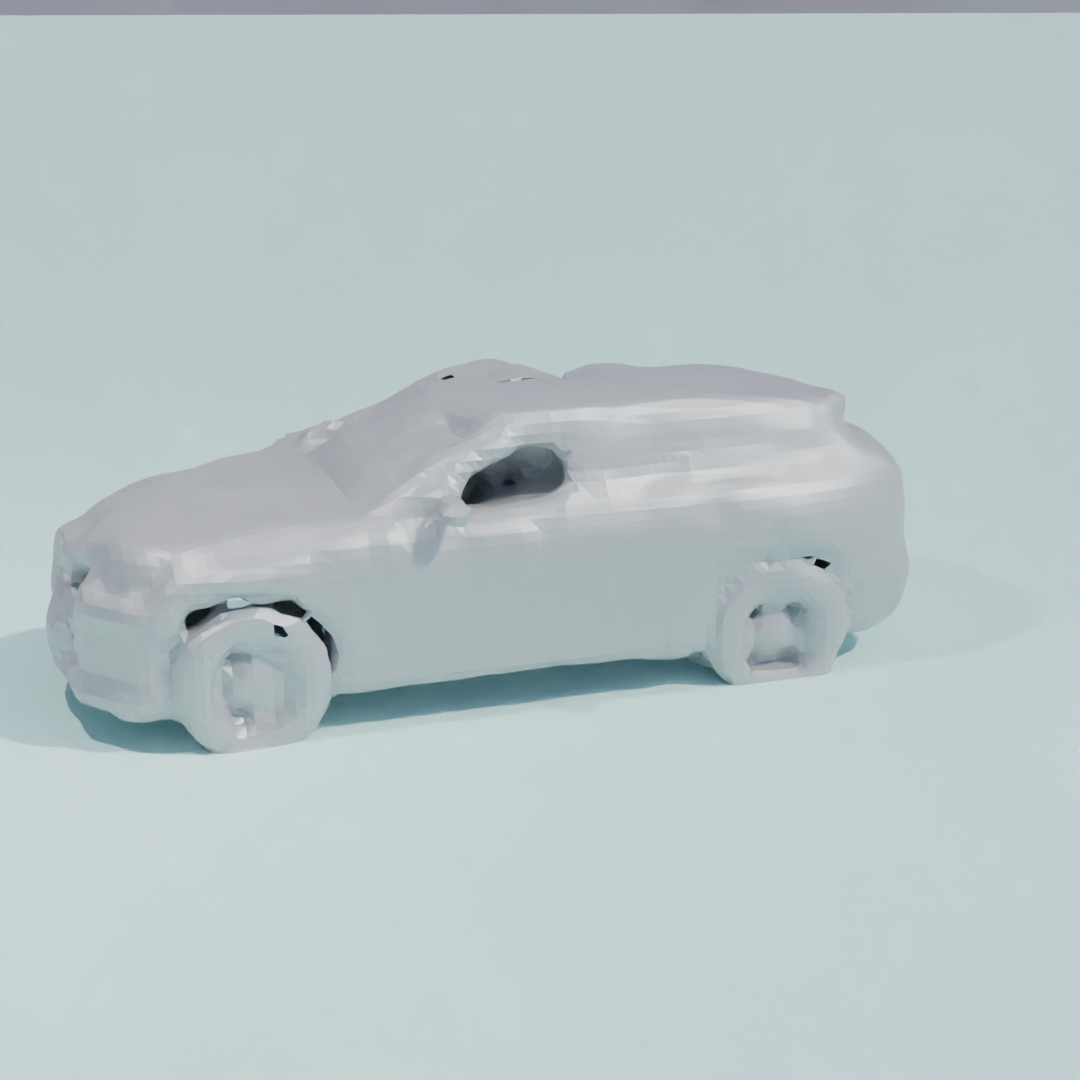}}\\\vspace{0.1cm}
    
    \frame{\includegraphics[width=0.22\textwidth,clip,trim = 0cm 7cm 0cm 9cm]{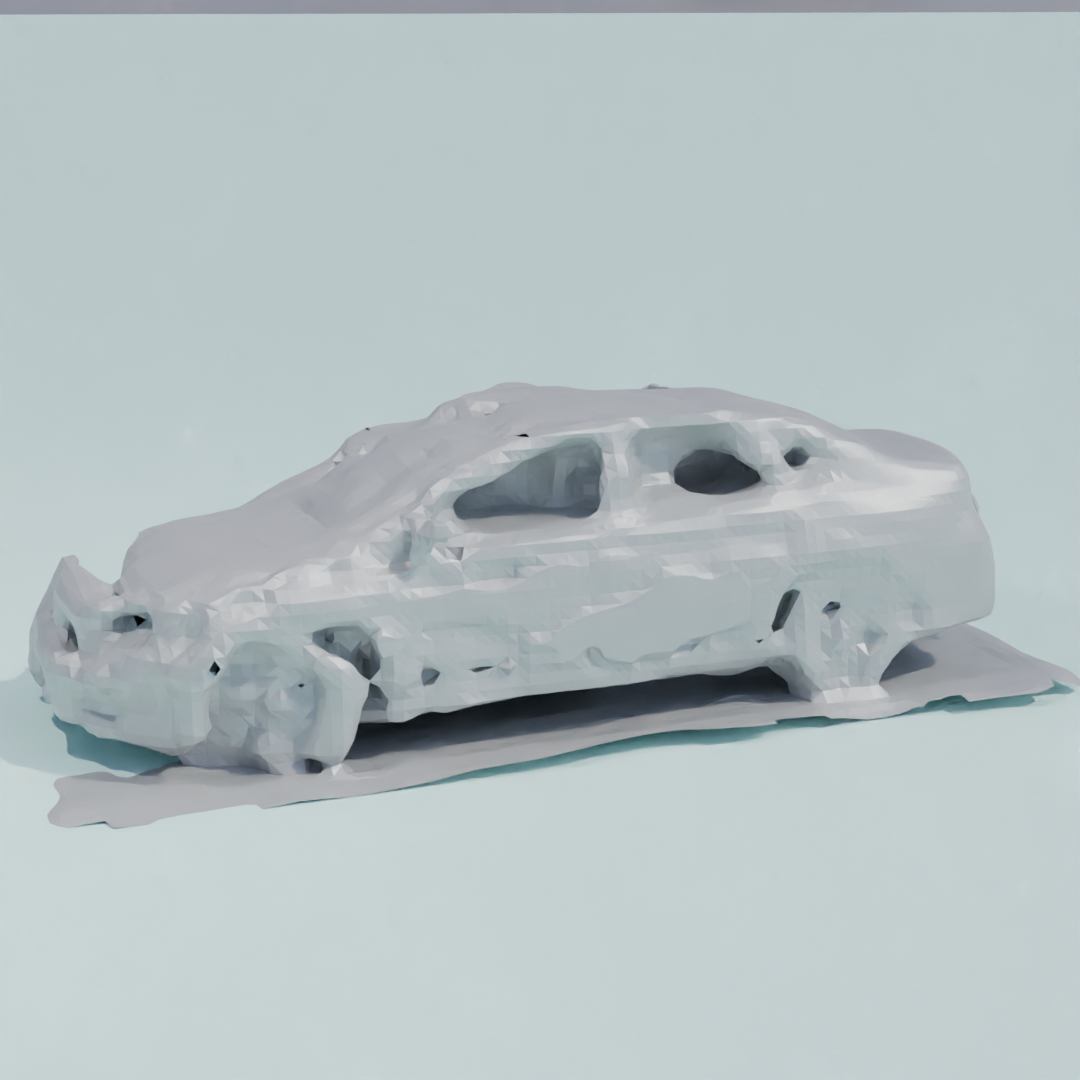}}\hspace{-0.05cm}
    \frame{\includegraphics[width=0.22\textwidth,clip,trim = 0cm 7cm 0cm 9cm]{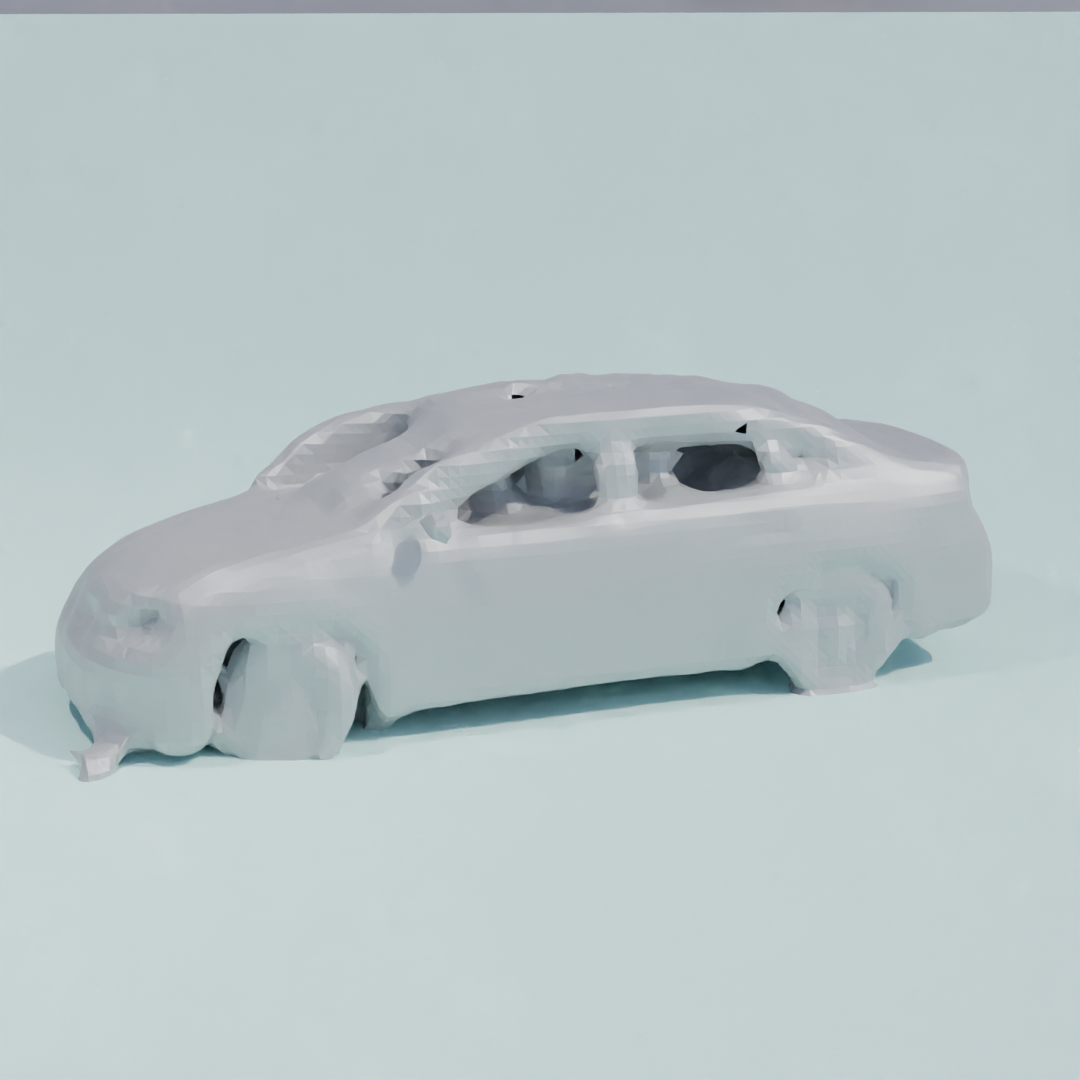}}\vspace{0.1cm}
    
    \frame{\includegraphics[width=0.22\textwidth,clip,trim = 0cm 9cm 0cm 10cm]{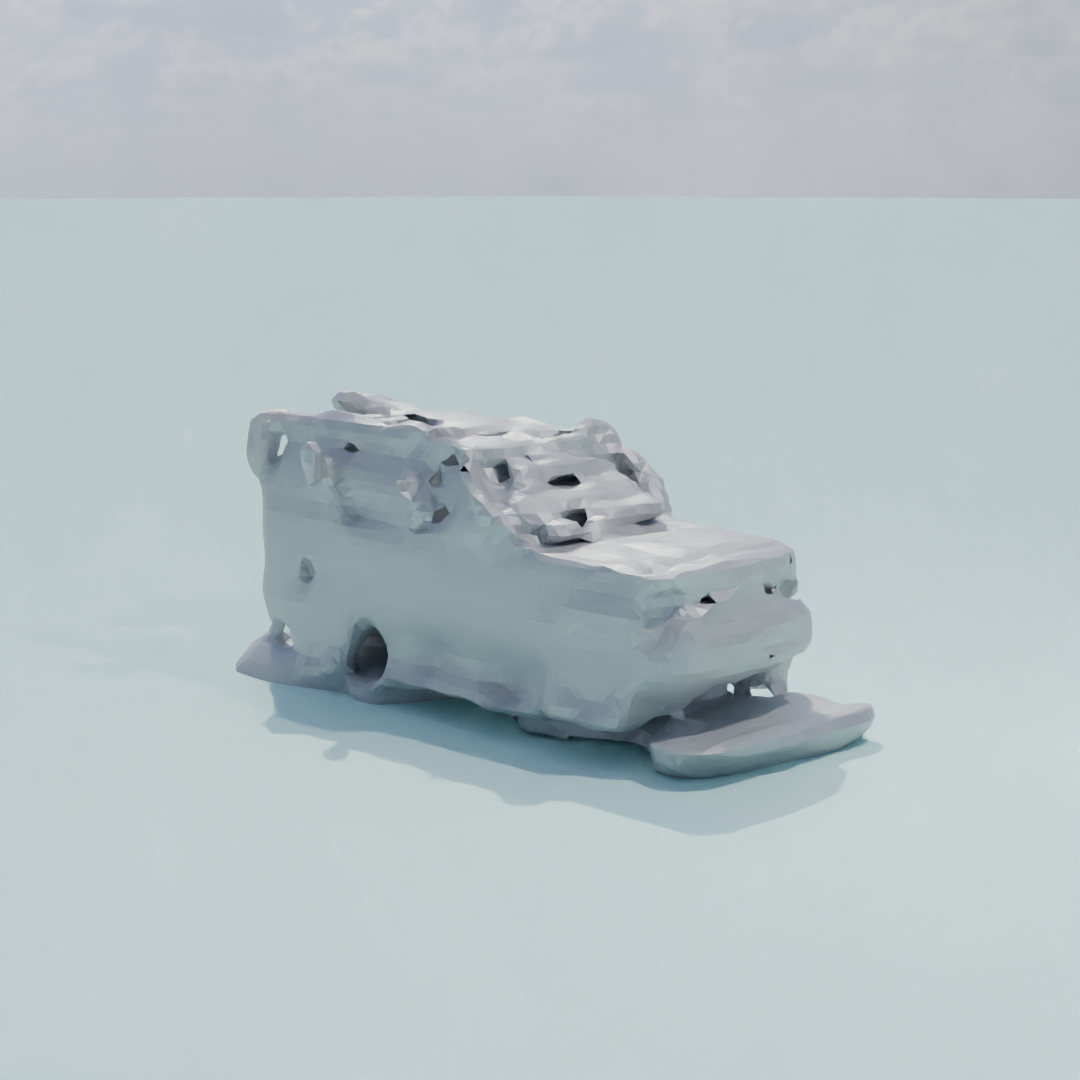}}\hspace{-0.05cm}
    \frame{\includegraphics[width=0.22\textwidth,clip,trim = 0cm 9cm 0cm 10cm]{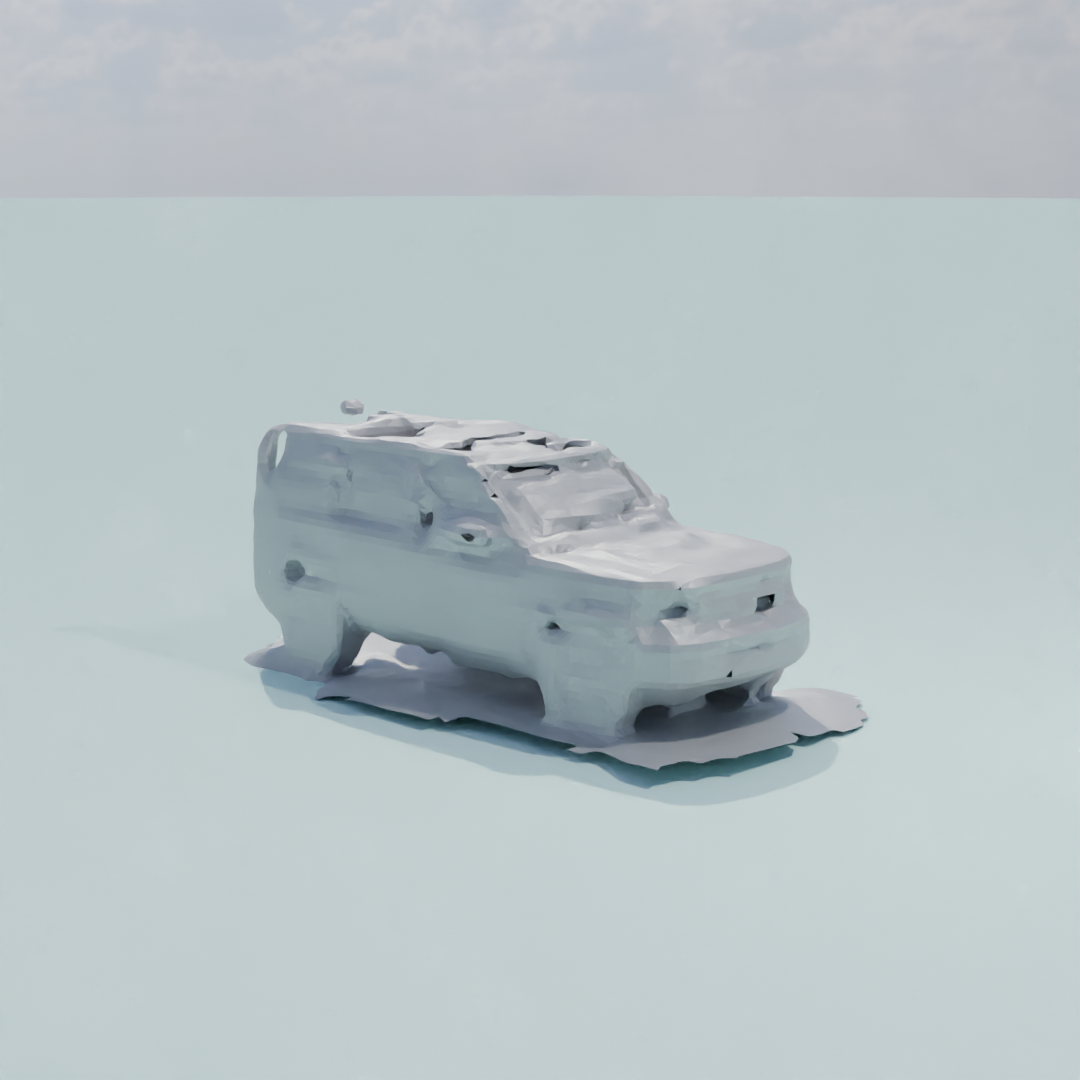}}\\\vspace{0.1cm}

    \frame{\includegraphics[width=0.22\textwidth,clip,trim = 0cm 7cm 0cm 8cm]{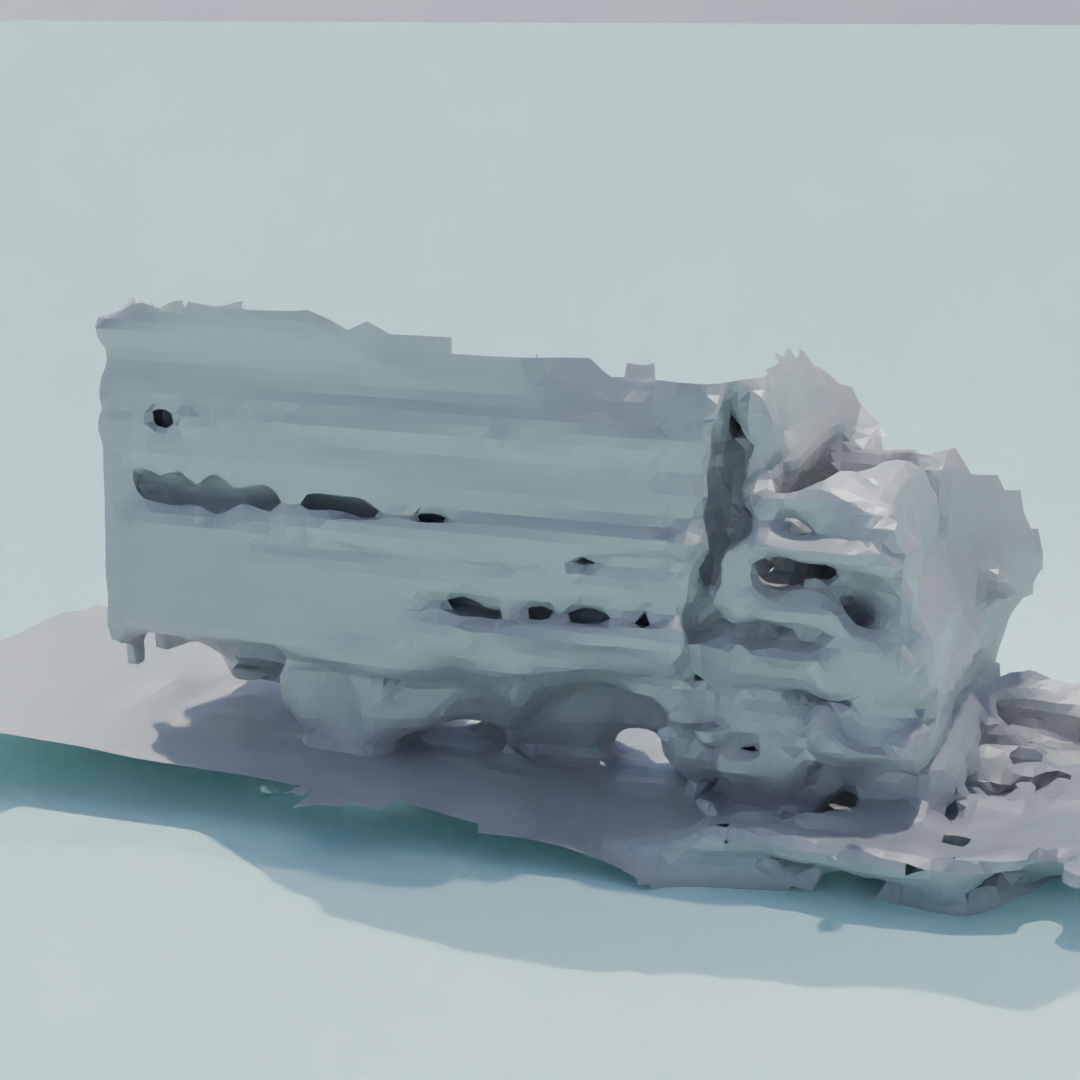}}\hspace{-0.05cm}
    \frame{\includegraphics[width=0.22\textwidth,clip,trim = 0cm 7cm 0cm 8cm]{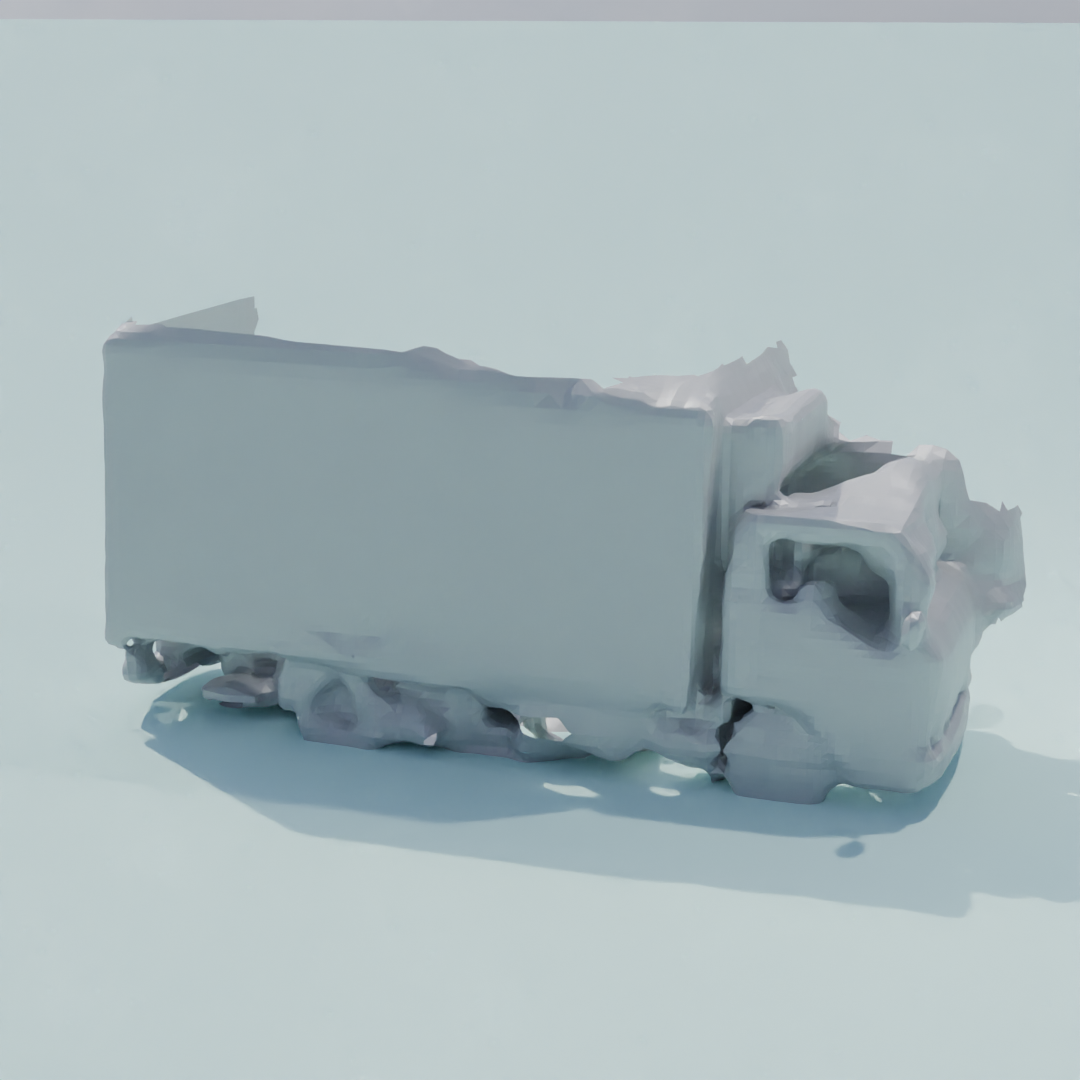}}\vspace{0.1cm}
    
    \caption{(\textbf{Left)} Each column shows nuScenes and Argoverse object reconstructions using ground truth poses compared to (\textbf{right}) ours.}\vspace{-0.3cm}
    \label{fig:nusc-fg}
\end{figure}

\section{Qualitative Results}
Visualizations of our foreground reconstructions on nuScenes and Argoverse are shown in \cref{fig:nusc-fg}. In the visualizations, we show that the ground-truth object annotations are not accurate enough to yield good reconstructions. Errors in bounding box alignment and orientation lead to point aggregation errors, leading to poor surface reconstructions. Note that this shows that the human-provided labels are not accurate enough for high-quality reconstruction. Motion distortion from dynamic objects also contributes to the poor quality of the ground truth reconstructions. In \cref{fig:ablation}, we show how accounting for this distortion can significantly improve the reconstructions.

\begin{figure}
    \centering
    \frame{\includegraphics[width=0.22\textwidth]{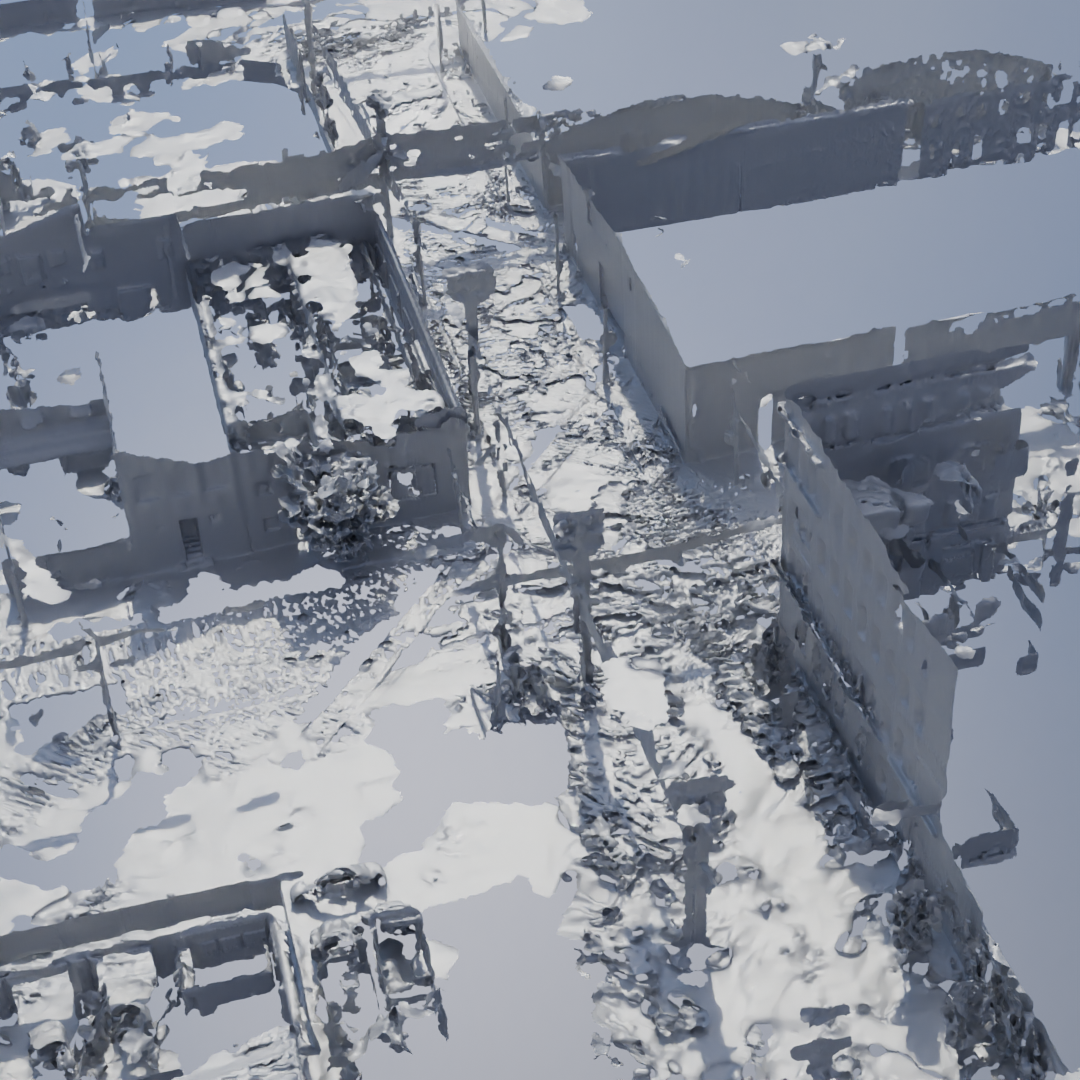}}\hspace{0.1cm}
    \frame{\includegraphics[width=0.22\textwidth]{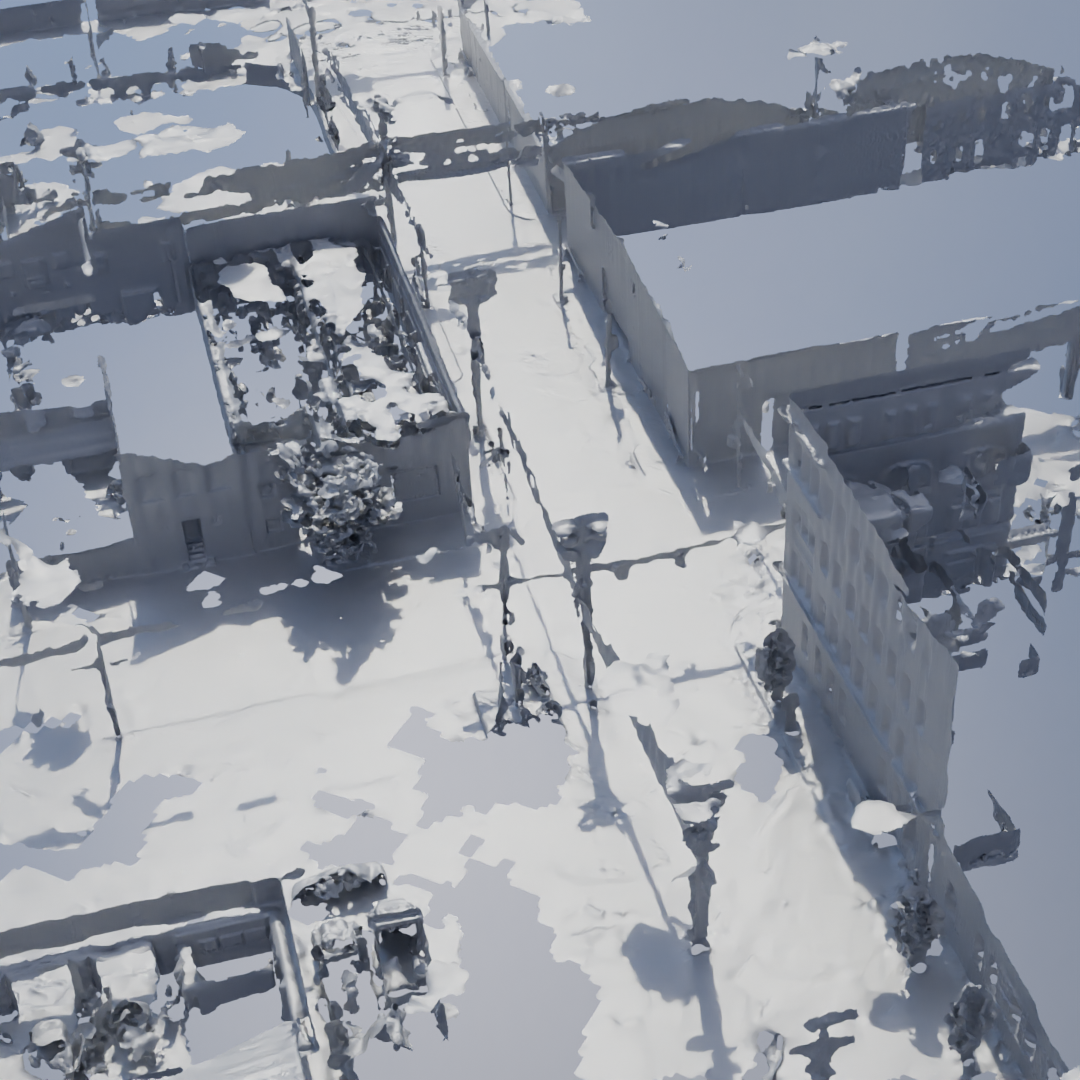}}\\\vspace{0.1cm}
    \frame{\includegraphics[width=0.22\textwidth,clip,trim= 0 0 0 2cm]{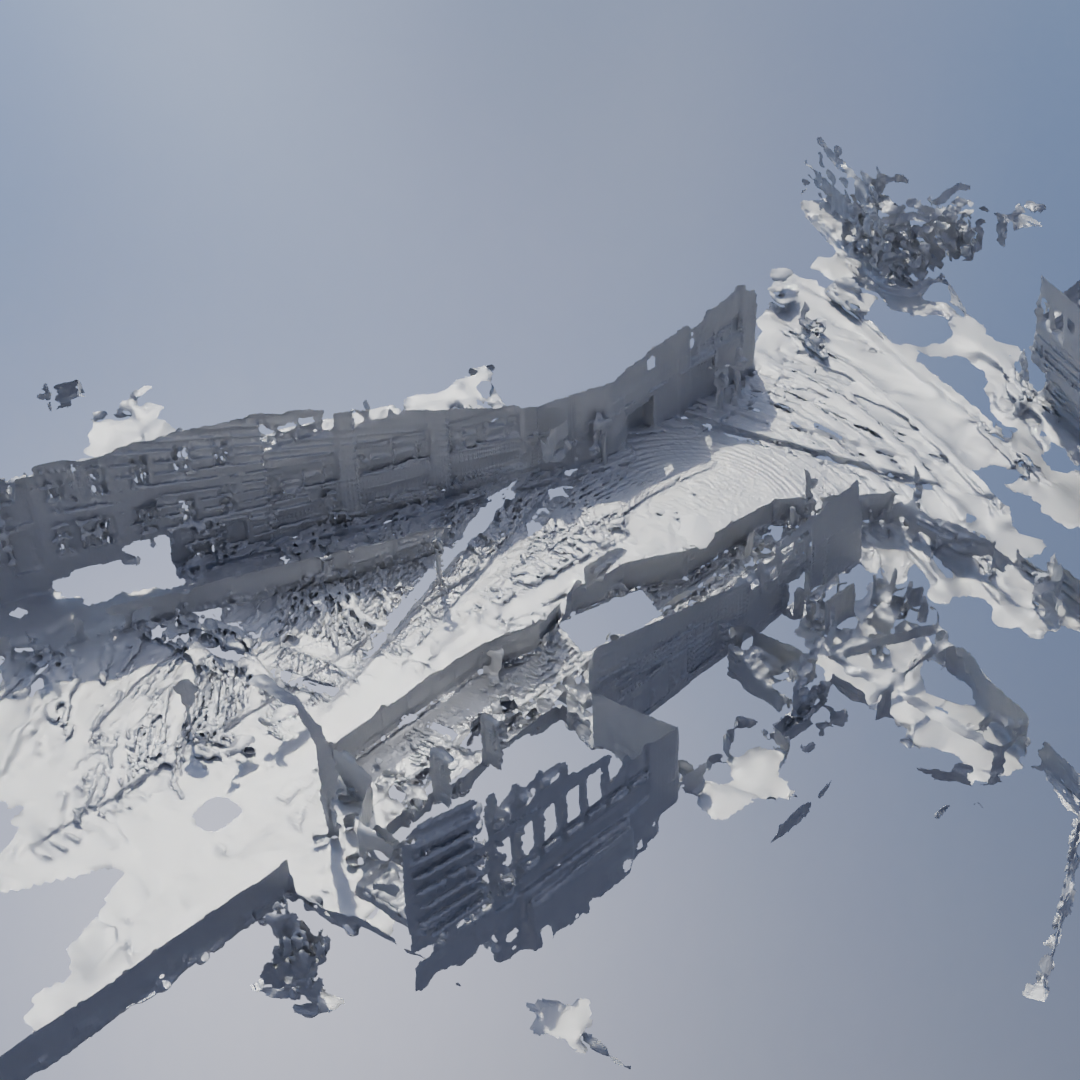}}\hspace{0.1cm}
    \frame{\includegraphics[width=0.22\textwidth,clip,trim= 0 0 0 2cm]{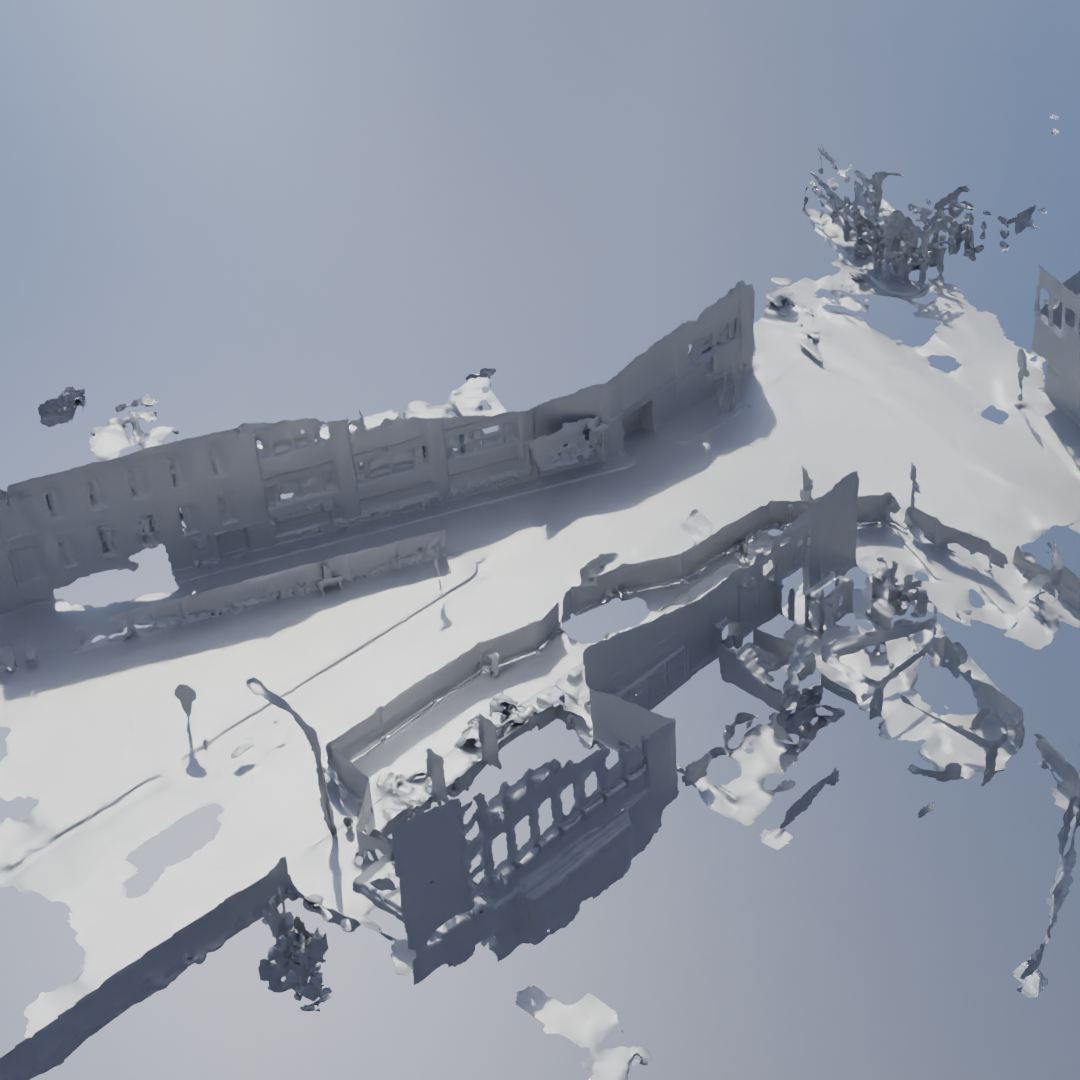}}
    \caption{Map reconstruction comparisons on Argoverse (\textbf{top}) and nuScenes (\textbf{bottom}). The \textbf{left} column uses poses from a high-quality LiDAR odometry method\cite{vizzo2023ral} and \textbf{right} shows the result from our method's poses. The ground truth poses are reasonably accurate, but small errors in orientation and position lead to artifacts on flat surfaces and missing small details.} 
    \label{fig:bg-av2}
\end{figure}

Background reconstructions from nuScenes and Argoverse are shown in \cref{fig:bg-av2}. For these ``objects'' the quality improvement comes from refining the ego-pose of the vehicle. As with foreground objects, ego-pose errors cause misalignment of the LiDAR sweeps, which become surface artifacts. However, the comparison is with a state-of-the-art LiDAR odometry method instead of the ground truth since we find odometry is generally superior.
\begin{table}
    \centering
    \small
    \begin{tabular}{lccc}
    \toprule
    & NN Dist (m) $\downarrow$ & Acc Relax $\uparrow$ & Acc Strict $\uparrow$\\
    \midrule
    NKSR\cite{huang2023nksr} + GT tracks\hspace{-0.5cm} & 0.071 & 0.9 & 0.76\\
    NKSR\cite{huang2023nksr} + LT3D\cite{peri2023towards}\hspace{-0.5cm} & 0.071 & 0.9 & 0.76\\
    \midrule
    Ours + GT tracks & 0.048 & 0.96 & 0.91\\
    Ours + GT tracks  (1 Hz)\hspace{-0.5cm} & 0.050 & 0.96 & 0.90\\
    Ours + GT tracks  (0.5 Hz)\hspace{-0.5cm} & 0.048  & 0.96  & 0.91\\
    Ours + GT tracks  (0.25 Hz)\hspace{-0.5cm} & 0.048 & 0.96 & 0.91\\
    Ours + LT3D\cite{peri2023towards} & 0.048 & 0.96 & 0.90\\
    \bottomrule
\end{tabular}
    \caption{Evaluation of our method's robustness to actor annotation errors (subsampling or real tracks). We measure reconstruction accuracy using the nearest-neighbor distance between the input point clouds and the reconstructed scene at each timestamp. We report the average distance and two accuracy metrics to characterize the distribution of errors. Specifically, we compute the percent of points less than 10cm and 5cm for the relaxed and strict metrics, respectively}
    \label{tab:surf}
\end{table}

\begin{table}
    \small
    \centering
    \begin{tabular}{lccc}
    \toprule
    Annotation Rate & 1Hz & 0.5Hz & 0.25Hz\\
    \midrule
    Interpolation & 0.29m & 0.40m & 0.74m\\
    Ours  & \textbf{0.20m} & \textbf{0.22m} & \textbf{0.52m}\\
    \bottomrule
\end{tabular}
    \caption{Pose accuracy evaluation on nuScenes (using NuScene's default ATE metric), measured by comparing the bounding box locations predicted by our method to held-out ground truth labels provided at 2Hz. We compare our method to linearly interpolating the poses as is commonly done to create scene-flow labels~\cite{baur2021slim}.}
    \label{tab:actor-pose}
\end{table}
\begin{table}
    \small
    \centering
\begin{tabular}{lcccc}
    \toprule
    Method & Chamfer Dist. $\downarrow $ &  Depth $\downarrow$\\
    \midrule
    Ours  & \textbf{0.26} & \textbf{0.0002} \\
    \midrule
    NeuRAD\cite{tonderski2024neurad} & 3.26 & 0.0053\\
    NeuRAD\cite{tonderski2024neurad} w. Our Poses & 2.19 & 0.002\\
    \bottomrule
\end{tabular}
    \caption{Our reconstructions sigificantly outperform the state-of-the-art in LiDAR Novel View Synthesis~\cite{tonderski2024neurad} in terms of chamfer distance ($m^2$) and median depth error ($m^2$). Such methods can benefit from our optimized poses, suggesting that SMORE produces more accurate poses and benefits from an explicit surface model (\cref{fig:pcd}).}
    \vspace{-0.3cm}
    \label{tab:nvs}
\end{table}
\begin{table}
    \small
    \centering
    \begin{tabular}{lccccc}
    \toprule
    Pose Source & PSNR $\uparrow$ & SSIM $\uparrow$ & LPIPS $\downarrow$ & CD $\downarrow$ & Depth $\downarrow$ \\
    \midrule
    Nuscenes-GT & 26.37 & 0.791 & 0.283 & 4.22 & 0.017 \\
    KISS-ICP\cite{vizzo2023ral} & 24.99 & 0.76 & 0.328 & 2.58 & 0.016 \\
    Ours & \textbf{27.52} & \textbf{0.821} & \textbf{0.238} & \textbf{2.19} & \textbf{0.002} \\
    \bottomrule
\end{tabular}
    \caption{Ego-pose evaluation by fitting a SOTA NeRF method~\cite{tonderski2024neurad} to: the poses provided by nuScenes, those from a LiDAR odometry method, and the ones produced by our method. We see that our poses yield a superior scene model.}
    \label{tab:ego-pose-neurad}
\end{table}
\section{Conclusion}
In this work, we brought dense, dynamic reconstruction to the large-scale in-the-wild autonomous vehicle setting. We developed an optimization framework for understanding this problem and provided a simple yet effective solution based on decomposing the problem into well-studied sub-components. This solution yields high-quality reconstructions of both the foreground and background and can even account for subtle distortions in the input point clouds. We hope that this method will not only be useful for creating training and evaluation data for other perception tasks but will also promote active research in this challenging setting.
\clearpage
\bibliographystyle{splncs04}
\bibliography{main}
\clearpage
\section{LiDAR Novel View Synthesis Experimental Details}
We extend the novel view synthesis comparison to more scenes in Nuscenes. In the main paper, we selected {\tt scene-0103} as the candidate choice of scene due to sensor pose variations along the z-axis. We show LiDAR NVS results on 3 more scenes from the mini-val dataset in Table~\ref{tab:nvs_full}. We compare SMORE against NeuRAD by considering 3 scenarios-  first: using NuScenes poses with no pose optimization, second: optimizing poses and third: directly using SMOR optimized poses. We observe that our method consistently outperforms NeuRAD by an order of magnitude on chamfer distance as well as median L2 depth across all scenes. 
\begin{table}[htb!]
    \centering
    \small
\begin{tabular}{lcccc}
    \toprule
    Method & Chamfer Dist. $\downarrow $ &  Depth $\downarrow$\\
    \midrule
    \textbf{scene-0061} & & \\
    \midrule
    NeuRAD\cite{tonderski2024neurad} w. Nuscenes Poses & 1.81 &	0.0081 \\
    NeuRAD\cite{tonderski2024neurad} w. Ego-pose Optimization& 2.87 &	0.0102 \\
    NeuRAD\cite{tonderski2024neurad} w. Our Poses & 1.98 &	0.0022\\
    Ours  & \textbf{0.36} & \textbf{0.0005} \\
    \midrule
    
    \textbf{scene-0103} & & \\
    \midrule
    NeuRAD\cite{tonderski2024neurad} w. Nuscenes Poses & 4.22	& 0.0170 \\
    NeuRAD\cite{tonderski2024neurad}  w. Ego-pose Optimization & 3.26 & 0.0053\\
    NeuRAD\cite{tonderski2024neurad} w. Our Poses & 2.19 & 0.0020\\
    Ours  & \textbf{0.26} & \textbf{0.0002} \\
    \midrule

    \textbf{scene-0796} & & \\
    \midrule
    NeuRAD\cite{tonderski2024neurad} w. Nuscenes Poses & 3.80 &	0.0546 \\
    NeuRAD\cite{tonderski2024neurad}  w. Ego-pose Optimization& 3.68 &	0.0245\\
    NeuRAD\cite{tonderski2024neurad} w. Our Poses & 3.60 & 	0.0177\\
    Ours  & \textbf{0.31} &	\textbf{0.0002} \\
    \midrule

    \textbf{scene-1094} & & \\
    \midrule
    NeuRAD\cite{tonderski2024neurad} w. Nuscenes Poses & 2.53	& 0.0566 \\
    NeuRAD\cite{tonderski2024neurad}  w. Ego-pose Optimization & 3.17 &	0.0452\\
    NeuRAD\cite{tonderski2024neurad} w. Our Poses & 2.41	& 0.0120\\
    Ours  & \textbf{0.43} & \textbf{0.0007} \\
    \bottomrule
\end{tabular}
    \caption{\textbf{LiDAR Novel View Synthesis on more scenes from NuScenes}. SMORE consistently outperforms NeuRAD across a variety of scenes by an order of magnitude.}
    \label{tab:nvs_full}
\end{table}

\section{Ego-Pose Evaluation}
We conduct experiments on more scenes to evaluate the efficacy of ego-poses recovered by SMORE's optimization. Similar to the main paper, we observe that poses generated by our method significantly outperform the NuScenes ground-truth as well as those from a LiDAR odometry~\cite{vizzo2023ral} across all scenes (See Table~\ref{tab:ego-pose-neurad-full}). 
\begin{table}[htb!]
    \centering
    \small
    \begin{tabular}{lccccc}
    \toprule
    Pose Source & PSNR $\uparrow$ & SSIM $\uparrow$ & LPIPS $\downarrow$ & CD $\downarrow$ & Depth $\downarrow$ \\
    \midrule
    
    \textbf{scene-0061} & & & & & \\
        \midrule 
        Nuscenes-GT & 24.93	& 0.752 &	0.357	& 1.81	& 0.0081 \\
        KISS-ICP\cite{vizzo2023ral} & 23.99	& 0.729	& 0.396	& 2.65	& 0.0139 \\
        Ours & \textbf{25.98} & 	\textbf{0.781 }&  \textbf{0.306}	& \textbf{1.98} & \textbf{0.0022}\\
        \midrule
    \textbf{scene-0103} & & & & & \\
        \midrule 
        Nuscenes-GT & 26.37 & 0.791 & 0.283 & 4.22 & 0.017 \\
        KISS-ICP\cite{vizzo2023ral} & 24.99 & 0.76 & 0.328 & 2.58 & 0.016 \\
        Ours & \textbf{27.52} & \textbf{0.821} & \textbf{0.238} & \textbf{2.19} & \textbf{0.002} \\
        \midrule
    \textbf{scene-0796} & & & & & \\
        \midrule 
        Nuscenes-GT & 22.25 & 	0.629	 & 0.528 & 	3.80	& 0.0546 \\
        KISS-ICP\cite{vizzo2023ral} & 21.63	& 0.622	& 0.544 &	3.95 &	0.0388 \\
        Ours & \textbf{22.44} & 	\textbf{0.647} & 	\textbf{0.514} & 	\textbf{3.60} &	\textbf{0.0177} \\
        \midrule
    \textbf{scene-1094} & & & & & \\
        \midrule 
        Nuscenes-GT & 23.59	& 0.512 & 	0.526 & 	2.53 &	0.0566 \\
        KISS-ICP\cite{vizzo2023ral} & 22.93	& 0.505	& 0.585	& 2.75	& 0.0501 \\
        Ours & \textbf{24.28} & 	\textbf{0.528} & 	\textbf{0.492}	& \textbf{2.41} & 	\textbf{0.0120} \\

    \bottomrule
\end{tabular}
    \caption{\textbf{Ego-pose evaluation by fitting NeuRAD on more scenes from NuScenes}. SMORE's refined poses provide improved geometry estimates when compared to other poses, including even the ground truth poses provided by NuScenes.}
    \label{tab:ego-pose-neurad-full}
\end{table}

\section{Bounding Box Evaluation Details}
We used the standard NuScenes object detection metric, average translation error, to measure our method's improvement of the ground-truth bounding boxes. However, as our method is not an object detector, that comparison has some complications, which we explain here.

The average translation error is defined as the distance between the centers of the predicted and ground truth bounding boxes. Since our method does not predict bounding boxes, first, we need to define a ``center'' for them. The center has no meaning for our reconstruction, so we can choose any fixed point on each object. Specifically, we choose the point that minimizes the sum of square distances to the centers of all the input bounding boxes. Note that when we subsample the inputs for evaluation, we do not use the held-out boxes to determine the ``predicted center''.

Next, we must define what constitutes a ``detection'' for our algorithm. Our reconstructions are formed by aggregating many points over multiple sweeps, which are registered to the predicted surfaces. Due to the labeling procedure of NuScenes, some of these input bounding boxes contain very few LiDAR returns ($<$ 50). The lack of points causes ambiguities in the registration step and can lead to instabilities, so we drop them from the optimization. In table \cref{tab:actor-pose}, we show the results on only boxes that have been optimized by our method.




\section{Argoverse 2.0 Evaluation}
\begin{table}
    \scriptsize
    \centering
    \begin{tabular}{lccc}
\footnotesize
    & NN Dist (m) $\downarrow$ & Acc Relax $\uparrow$ & Acc Strict $\uparrow$\\
    \midrule
    NKSR\cite{huang2023nksr} + GT tracks (10Hz) + GT ego-pose& 0.086 & 0.89 & 0.76\\
    NKSR\cite{huang2023nksr} + LT3D\cite{peri2023towards} tracks  + GT ego-pose & 0.088 & 0.89 & 0.73\\
    \midrule
    Ours + GT tracks (10 Hz) + KISS ego-pose\cite{vizzo2023ral} & 0.074 & 0.93 & 0.82\\
    Ours + GT tracks (10 Hz) + GT ego-pose & 0.079 & 0.93 & 0.81\\
    Ours + GT tracks  (5 Hz) + GT ego-pose & 0.073 & 0.93 & 0.83\\
    Ours + GT tracks  (2.5 Hz) + GT ego-pose & 0.073 & 0.93 & 0.83\\
    Ours + LT3D\cite{peri2023towards} tracks + GT ego-pose & 0.083 & 0.92 & 0.79\\
    \bottomrule
\end{tabular}\vspace{0.1cm}
    \caption{Surface quality evaluation on Argoverse 2.0, measured by comparing the LiDAR points to their closest points on the reconstructed surfaces.}
    \label{tab:surf-av2}
\end{table}
We replicated the same robustness evaluation done on NuScenes on Argoverse 2.0. As with NuScenes we use a small subset of the validation dataset for our evaluation. Specifically, sequences: a7636fca-4d9e-3052-bef2-af0ce5d1df74, 0c3bad78-9f1e-395d-a376-2eb7499229fd, e50e7698-de3d-355f-aca2-eddd09c09533, 0aa4e8f5-2f9a-39a1-8f80-c2fdde4405a2 d770f926-bca8-31de-9790-73fbb7b6a890.

As with NuScenes, we tested our method with various modifications to the inputs, either downsampling the ground truth annotations or by using tracked produced by LT3d\cite{peri2023towards}. The results can be found in tables \cref{tab:surf-av2} and reconfirm our main findings in the NuScenes results: our method can produce high-quality reconstructions even with input annotations of significantly worse quality than the ground truth. Again, we see a significant improvement over reconstructions using the ground truth poses.
\section{Failure Cases}

\textbf{Ground Holes in AV2 Background Reconstructions:} We find (and show in \cref{fig:bg-av2}) that the ground surface we extract from Argoverse is not as complete as those we extract from NuScenes. We believe that this is the result of the orientation of the LiDAR lasers used in each dataset collection. The LiDAR lasers in AV2 are oriented such that they focus the resolution "down-range" to make detecting vehicles and pedestrians easier. This results in less resolution on the ground. To see this, compare the distance between laser returns near the car in NuScenes and Argoverse in \cref{fig:ray-vis}. Despite this, we still believe that good reconstructions of the ground should be possible and investigating this is an area of future research.

\textbf{Registration Failures:} Another source of errors for our method is when ICP produces a poor registration on a vehicle. One common cause of this is attempting to register a sweep to a vehicle that contains very few points. We use a heuristic to filter out most of these cases (dropping views of an object with fewer than 50 points) but some can still cause errors which manifest as "jittery" motion of objects. Another, harder to filter, source of error is from registering scans with low "texture". In the context of ICP, low texture means scans which do not contain corners or edges useful for exact alignment. This can occur when only the side face of a vehicle is observed, resulting in a flat plane of points which has many possible alignments to the reconstructed shape. We believe that both of these errors can be mitigated by applying stronger motion priors to the reconstructed objects in order to add constraints to the system. This is another direction for future work.

\end{document}